\documentclass[sn-mathphys]{sn-jnl}

\usepackage{amssymb}
\usepackage{amsthm}
\usepackage{amsmath}
\usepackage{subcaption}
\usepackage{multirow}
\usepackage{fancyhdr}
\usepackage{array}
\usepackage{booktabs}
\usepackage{float}
\usepackage{epstopdf}
\usepackage{xcolor}
\newcolumntype{M}[1]{>{\centering\arraybackslash}m{#1}}

\jyear{2022}%

\theoremstyle{thmstyleone}%
\theoremstyle{thmstyletwo}%

\theoremstyle{thmstylethree}%
\newtheorem{definition}{Definition}%

\begin{document}

\title[Spatial-Temporal Meta-path Guided Explainable Crime Prediction]{Spatial-Temporal Meta-path Guided Explainable Crime Prediction}


\author[1]{\fnm{Yuting} \sur{Sun}}\email{yuting.sun@uqconnect.edu.au}

\author[1]{\fnm{Tong} \sur{Chen}}\email{tong.chen@uq.edu.au}

\author*[1]{\fnm{Hongzhi} \sur{Yin}}\email{db.hongzhi@gmail.com}

\affil[1]{\orgdiv{School of Information Technology and Electrical Engineering}, \orgname{The University of Queensland}, \orgaddress{
\country{Australia}}}


\abstract{Exposure to crime and violence can harm individuals' quality of life and the economic growth of communities. In light of the rapid development in machine learning, there is a rise in the need to explore automated solutions to crime prevention. The increasing availability of both fine-grained urban and public service data has driven a recent surge in fusing such cross-domain information to facilitate crime prediction. By capturing the information about social structure, environment, and crime trends, existing machine learning predictive models have explored the dynamic crime patterns from different views. However, these approaches mostly convert such multi-source knowledge into implicit and latent representations (e.g., learned embeddings of districts), making it still a challenge to investigate the impacts of explicit factors for the occurrences of crimes behind the scenes. In this paper, we present a \textbf{\underline{S}}patial-\textbf{\underline{T}}emporal \textbf{\underline{M}}eta-path guided \textbf{\underline{E}}xplainable \textbf{\underline{C}}rime prediction (\textbf{STMEC}) framework to capture dynamic patterns of crime behaviours and explicitly characterize how the environmental and social factors mutually interact to produce the forecasts. Extensive experiments show the superiority of STMEC compared with other advanced spatial-temporal models, especially in predicting felonies (e.g., robberies and assaults with dangerous weapons).}




\maketitle

\section{Introduction}\label{sec1}

Crime is an inevitable and persistent problem that brings negative outcomes to society. It is reported that worldwide homicide has caused more than 400,000 deaths each year. Given 80\% of the victims are younger than 50 years old, homicide become one of the leading causes of death among young adults \cite{Abril07}. Apart from the long-lasting physical and psychological injuries the victims may suffer from, crime can also increase government expenditures on police protection and justice services. According to the report from the Australian Institute of Criminology, serious and organised crime has cost the Australian government up to 47.4 billion dollars in 2017 \cite{Harel78}. To minimize the effect on public safety and urban sustainability, rapid response policing is always required when government agencies are alerted on any criminal activities. In this case, crime prediction plays a key role in changing the situation from being blindsided to being better prepared, and is formulated as the task of predicting region-wise crime rates using historical records.

As crime data is a type of spatial-temporal event data, some effective data mining techniques have been proposed to explore the spatial and temporal features of crimes.
Temporal information in historical crime records concerning periodicity is emphasized by the routine activity theory \cite{Silas1}. 
Traditional time series models, such as Autoregression Integrated Moving Average (ARIMA) \cite{8530836}, Linear Regression and Support Vector Regression (SVR) \cite{9573225} are widely used to make short-term crime forecasting. However, these models emphasize more on recent data and assume a fixed pattern of seasonality limited by linear models. Unlike typical time series such as traffic and weather, crime events have more irregularity in its temporal patterns, restricting the effectiveness of these conventional methods. Additionally, such methods fall short when capturing the spatial connections among regions, which are crucial as a region's crime patterns can also be inferred via geographically adjacent \cite{10.1145/3132847.3133024} and structurally similar \cite{8594999} regions. Though methods based on recurrent neural networks (RNNs) \cite{10.1145/3269206.3271793, 10.1145/3308558.3313730} are proposed for crime prediction, they overlook the correlations between regions with either geographical \cite{10.1145/3269206.3271793} or semantic \cite{10.1145/3308558.3313730} (e.g., demographic structure) affinity. Thus, a model specifying the spatial-temporal dynamics of criminal behaviors is highly desired.

Based on the nature of crimes \cite{Thom1}, both social characteristics and geographical locations of the residential places exhibit strong correlations with crimes. With the increasing availability of data collected from different channels, it has recently become possible to fuse multi-source information to facilitate crime prediction \cite{10.1145/3394486.3403201}. For example, \cite{Wang2020DeepTM} aggregates various resources by formulating them as graph-structured data and merging the learned graph representations. However, by merely treating the auxiliary information as the input feature for each region, it fails to fully account for the impact of each factor within the fine-grained urban data (e.g., income level and demographic distribution), leading to a substantial loss of context information. Meanwhile, it is non-trivial to fuse and represent the multi-faceted information in a more expressive way to quantify their varying contributions to different crime events.

The advances in latent factor models, especially neural networks, have witnessed dominating performance in predictive tasks where crime prediction is no exception. Those methods normally represent regions as implicit embedding through aggregating the information in both spatial and temporal domains \cite{10.1145/3269206.3271793, 8594999, ijgi7080298, 10.1145/3308558.3313730}. Despite the capability of forecasting the occurrence of crime events by capturing temporal patterns and neighborhood characteristics, the rationale behind the resulted predictions largely remain unexplainable within most methods \cite{ 10.1145/3308558.3313730, 10.1145/3269206.3271793}. Understanding these factors that limit the communities from preventing crimes is of great importance to government agencies, as more effective measures can be taken to reduce crimes. While criminologists put a lot of effort into exploring reasons that cause crimes, the factors leading to criminal behaviors are still nowhere near well-understood \cite{don1}. Some methods with attention mechanisms \cite{10.1145/3269206.3271793} may distinguish the spatial and temporal effect on crime prediction, but the analysis on more specific factors (e.g., income level and population distribution) is left untouched, hindering both model performance and explanation quality. Apart from this, the importance of assumed factors may vary concerning different regions at different times. For example, even though economic and social disadvantages are believed to render areas crime-prone \cite{don1}, rural areas with a lower level of urbanization are less likely to attract offenders \cite{Edward1}.
However, most existing crime prediction methods do not differentiate such multifaceted information or further discuss the impact of features from multiple views \cite{10.1145/3308558.3313730, Wang2020DeepTM, Fateha01, ijdkp.2015.5401}, while statistical learning methods (e.g., linear regression and decision trees) reap direct interpretability but sacrifice the prediction accuracy \cite{j.physa.2018.03.084, Ginger17}.

To this end, we aim to predict the occurrence of crimes in each region by addressing the challenges in: (1) modeling the spatial-temporal effect; (2) expressively fusing multi-view information; and (3) providing interpretability on the causes of different crime types across regions. To address these challenges, we propose a novel framework - \textbf{\underline{S}}patial-\textbf{\underline{T}}emporal \textbf{\underline{M}}eta-path guided \textbf{\underline{E}}xplainable \textbf{\underline{C}}rime prediction (\textbf{STMEC}). Different from existing graph based and grid based crime prediction models \cite{10.1145/3269206.3271793, 8594999, ijgi7080298, 10.1145/3308558.3313730, Wang2020DeepTM} that capture the relationships between regions through either geographical distance or venue based similarity, STMEC can: (1) encode temporal dynamics and semantic similarity among regions into their representations to improve performance; (2) project multifaceted dimensions of auxiliary data into a heterogeneous graph to capture rich context information; and (3) explicitly model the interactions between regions and features into a distribution-aware path to enhance explainability.

It is worth noting that STMEC novelly depicts the relations between regions by different social and environmental factors, and its expected performance and explainability are strengthened via the path-enriched features in the graph-structured data. The idea is inspired by the findings from prior works mentioned above, which suggest that similar urban features or adjacent locations of regions will lead to similar crime patterns \cite{10.1145/3132847.3133024, 8594999}. To complement the problem of fusing multifaceted information in our model, we characterize the interactions between regions and features by leveraging the concept of the meta-path \cite{10.14778/3402707.3402736}. Compared with existing crime prediction methods \cite{10.1145/3269206.3271793, 10.1145/3308558.3313730,Wang2020DeepTM} that merely treat multi-view information as regions' features, we explicitly model such information into a heterogeneous graph, enabling us to mine different semantics and preserve the heterogeneity of information. {This meta-path based graph structure can benefit the crime prediction tasks by improving both model interpretability and predictive performance. On the one hand, as meta-path can explicitly capture semantic relations between regions and complex context features, it can help us make informed decisions and improve the interpretability of the model. On the other hand, compared with random walk based approaches \cite{hussein2018meta, chen2021uniting}, symmetric meta-paths can better capture region-wise similarity conditioned on each specific socioeconomic factor. Besides, compared with approaches that build relation-specific homogeneous graphs \cite{wang2021gsnet}, the meta-path based framework is more scalable, and is less likely to suffer from the sparsity of observed links between regions.} Also, to better capture spatial-temporal dynamics, we cast the meta-path based graph into multiple snapshots, where each snapshot contains learned temporal features from the historical criminal activities indicating the temporal dynamics. By distilling information from the region-to-region paths in each snapshot, we can further combine features from diverse paths.

Furthermore, with the help of the path based attention mechanism, the proposed STMEC framework can obtain weights that suggest the contribution of various paths and further enhance the interpretability of the model. The rationale of designing such meta-path schemas is that the criminal activities of a certain region can be inferred from the region with similar environmental and social structures or from its neighborhoods. For example, when predicting crimes in a downtown area with a higher income level and more diverse populations, apart from the local crime trends, we also believe that another region with similar income level and demographic distribution may share its crime patterns to some extent. In addition, a region is more likely to be crime-prone if its nearby regions are at a higher risk of attracting offenders. Hence, by fusing rich information from the knowledge based graph across the time slots, the framework is able to capture temporal dynamics and retrieve semantics from diverse paths in comparison with existing works.

In summary, we highlight the main contributions of our work as follows:
\begin{itemize}
\item We design a new multi-view crime prediction framework STMEC that can expressively capture complex spatial-temporal dependencies, as well as correlations with external factors.
\item We explicitly model the interactions between regions and features over time to benefit both the performance and explainability. Furthermore, this new perspective enriches the semantic context within the representations from the graph-structured crime data.
\item Extensive evaluations on two real-world datasets collected from New York City (NYC) have been performed, and the results show that STMEC surpasses state-of-the-art baselines in terms of both effectiveness and explainability.
\end{itemize}

\section{Preliminaries}\label{sec2}
\begin{figure*}[ht]
    \centering
     \begin{subfigure}[b]{0.3\textwidth}
         \centering
         \includegraphics[width=\textwidth]{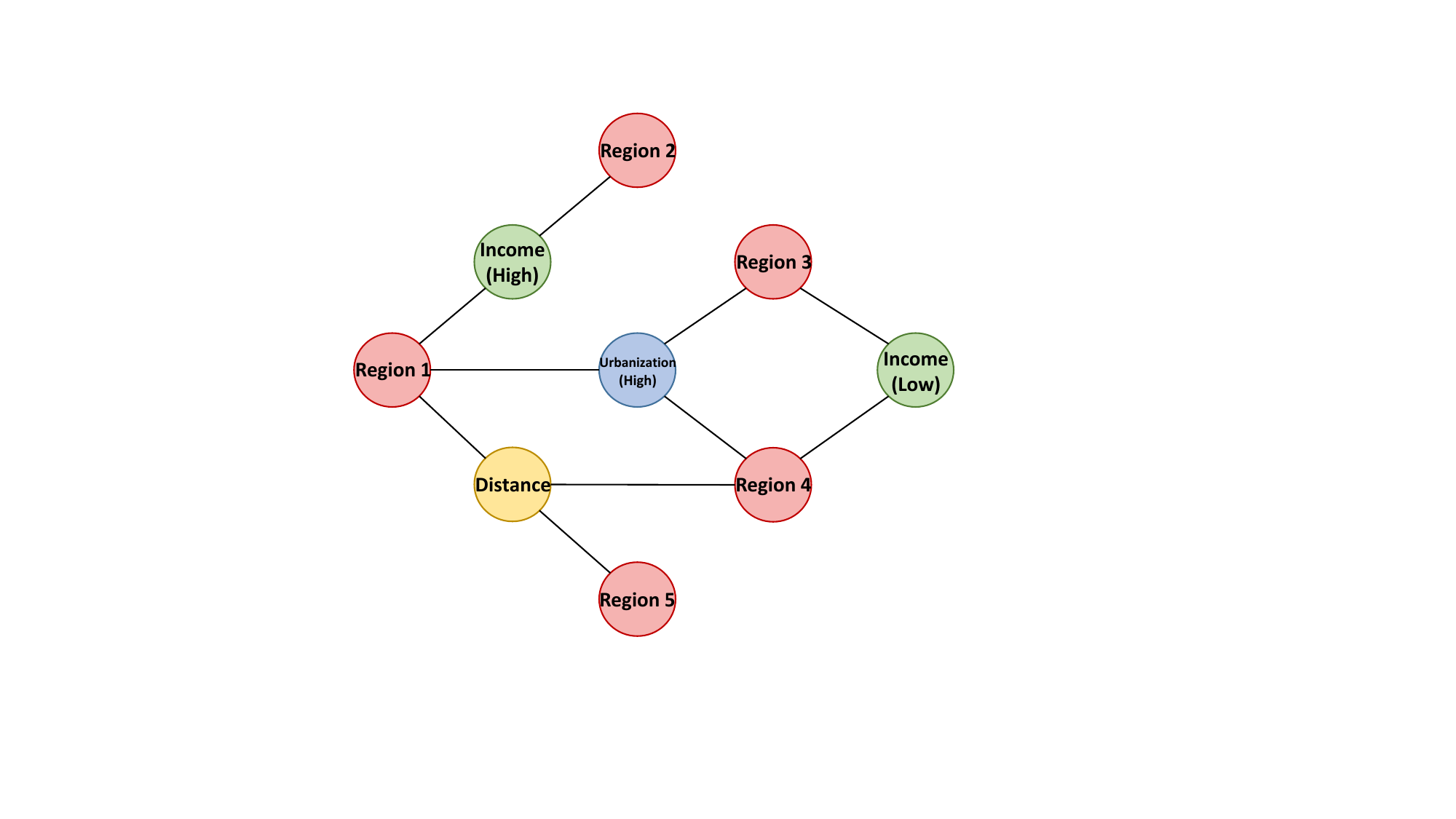}
         \caption{Heterogeneous Graph}
         \label{fig:heterogeneous}
     \end{subfigure}
     \hfill
     \begin{subfigure}[b]{0.24\textwidth}
         \centering
         \includegraphics[width=0.9\textwidth]{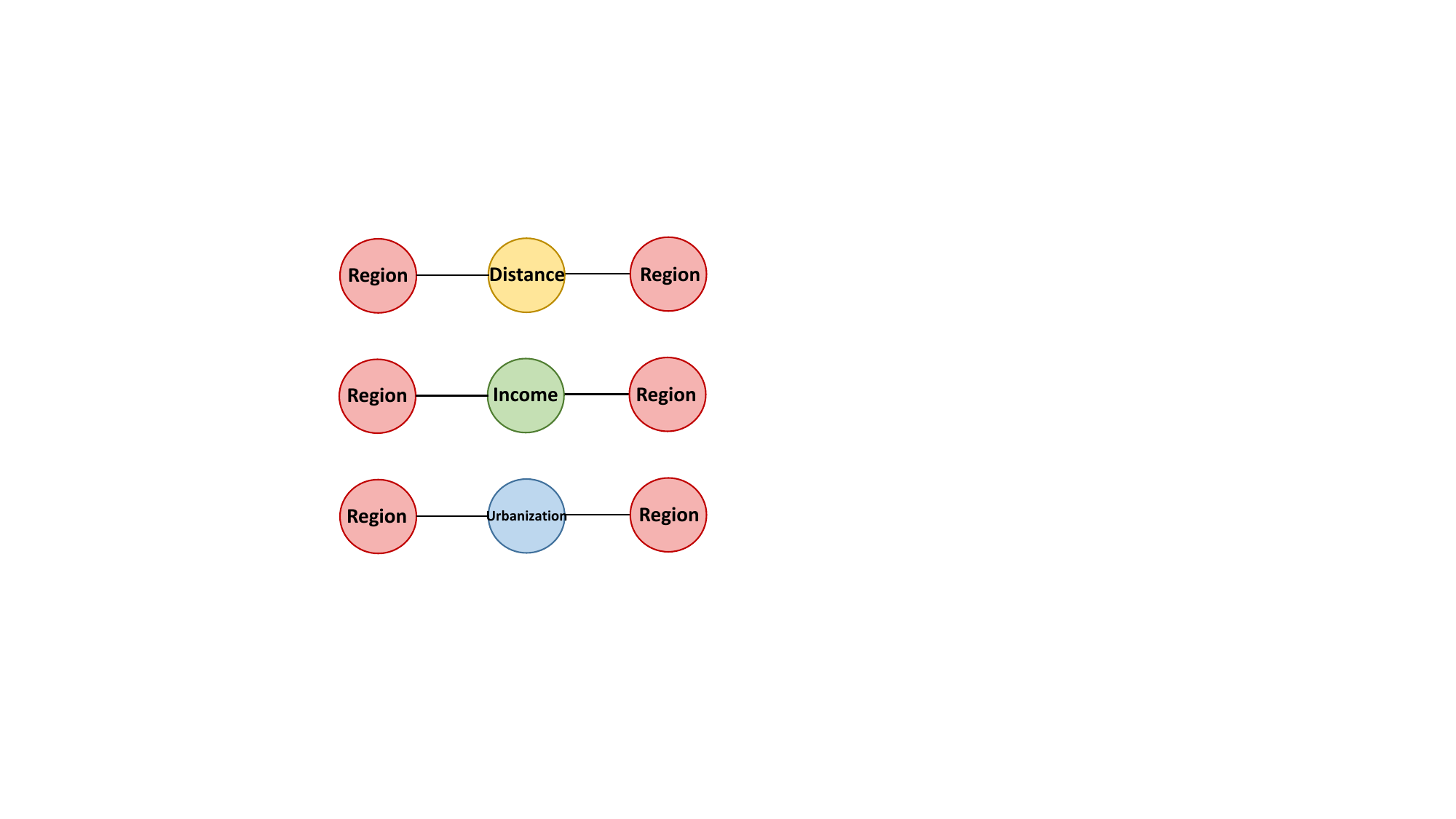}
         \caption{Meta-path Type}
         \label{fig:type}
     \end{subfigure}
     \hfill
     \begin{subfigure}[b]{0.43\textwidth}
         \centering
         \includegraphics[width=\textwidth]{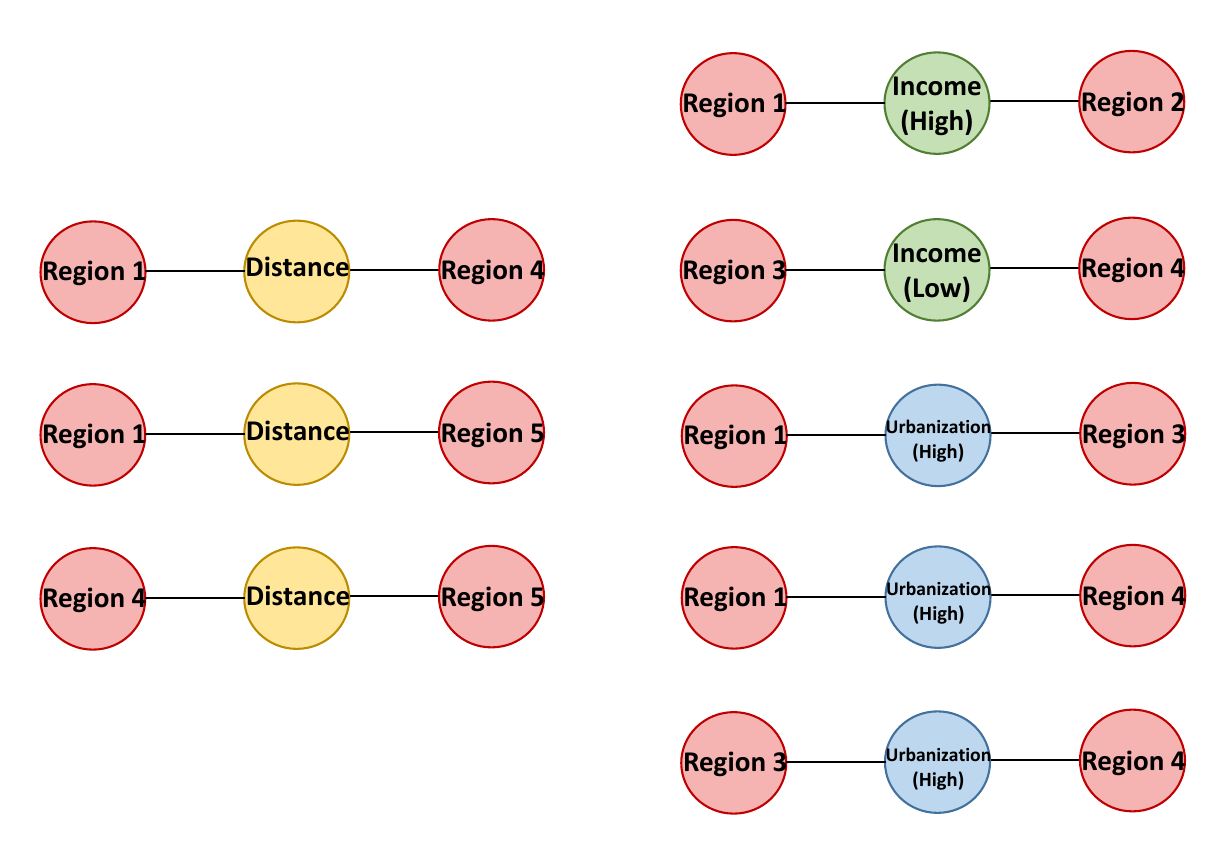}
         \caption{Meta-path Instances}
         \label{fig:instances}
     \end{subfigure}
        \caption{An illustration of meta-path based crime network. (a) An example of heterogeneous graph consisting of four types of nodes: region, income, urbanization, distance. (b) Three meta-path types in the heterogeneous graph denoted as $\langle region, factor, region \rangle$: $\langle region, income, region \rangle$ (RIR), $\langle region, urbanization, region \langle$ (RUR), $\langle region, distance, region \rangle$ (RDR). (c) Based on the three meta-path types, all 8 meta-path instances were found from the heterogeneous graph respectively.}
        \label{fig:example}
\end{figure*}

In this section, we begin with some essential definitions and present the problem formulation of crime prediction.
\begin{definition}
\textbf{Heterogeneous Information Network}. A heterogeneous information network (HIN) is defined as a graph $G=(V, E)$ with multiple entity types and relation types. Each entity $v \in V$ belongs to an entity type $\mathcal{A}$ given by the entity type mapping function $\phi: V \rightarrow \mathcal{A}$, and each edge $e \in E$ belongs to a relation type $\mathcal{R}$ given by the relation type mapping function $\psi: E \rightarrow \mathcal{R}$. By definition we have $\lvert\mathcal{A}\rvert+\lvert\mathcal{R}\rvert>2$.

In this work, there are 7 entity types representing regions, demographic, income level, job type, commuting ways, urban facility distribution, and geographic information. Given the sandwich structure of meta-path, which is denoted as $\langle region, factor, region \rangle$ in Figure \ref{fig:example}, the relation between region and factor is defined as `region contains information about factor'. In this work, the HIN is different from the pure region-only graphs, as the entity set contains both regions and attributes (e.g., income level and urban facilities).
\end{definition}

\begin{definition}
\textbf {Meta-path}. A meta-path $P$ in a network is denoted in the form of $A_{1} \stackrel{R_{1}}{\rightarrow} A_{2} \stackrel{R_{2}}{\rightarrow} \ldots \stackrel{R_{l}}{\rightarrow} A_{l+1}$. The composite relation from entity types $A_{1}$ to $A_{l+1}$ is described as $R=R_{1} \circ R_{2} \ldots \circ R_{l}$, where $A_{i} \in \mathcal{A}, R_{j} \in \mathcal{R}$.
\end{definition}

\begin{definition}
\textbf{Meta-path instances}. Given a meta-path $P$ of a heterogeneous graph $G$, a meta-path instance $p \in P$ describes a node sequence from entity $v_{0}$ to $v_{k}$ as $p=\langle v_{0}, v_{1}, \ldots v_{k}\rangle$, where $\forall i, \phi\left(v_{i}\right)=A_{i} \in \mathcal{A}$, and for each relation $e_{i}= \langle v_{i}, v_{i+1}\rangle, \psi\left(e_{i}\right)=R_{i} \in \mathcal{R}$.
\end{definition}

\textbf{Crime prediction}. For each geographic region in a city, we use $\mathcal{Y}_{i}=\{\mathbf{y}_{i}^{1}, \mathbf{y}_{i}^{2}, \ldots, \mathbf{y}_{i}^{t}, \ldots \mathbf{y}_{i}^{T}\} \in \mathbb{R}^{T \times C}$ to denote the occurrences of all $C$ crime types (1 for observed and 0 for unobserved) at region $r_{i}$ during $T$ time slots. Given the region set $R$, the previous crime records, and the region associated information network $G$, the objective of this work is to learn a predictive framework which infers whether a certain type of criminal activity will happen in the next time step $T+1$ at each region $r_{i} \in I$.

\begin{table}[ht]
  \small
  \centering
  \begin{tabular}{ccl}
    \toprule
    Notations&Definitions\\
    \midrule
    $G$ & Heterogeneous information network (HIN)\\
    $V$ & The set of nodes in a graph\\
    $E$ & The set of edges in a graph\\
    $v$ & A node or entity $v \in V$\\
    $e$ & An edge or relation $e \in E$\\
    $\mathcal{A}$ & Type of nodes (entities)\\
    $\mathcal{R}$ & Type of edges (relations)\\
    $P$ & A meta-path\\
    $p$ & A meta-path instance $p \in P$\\
    $I$ & A set of regions\\
    $r$  & A region $r\in I$\\
    $\mathbf{W}$ & Weight matrix\\
    $\mathbf{b}$ & Bias vector\\
    $\mathbf{h}$ & Embedding or hidden state\\
  \bottomrule
\end{tabular}
\caption{Description of notations used in this paper.}
 \label{tab:notations}
\end{table}

The key notations used in this paper are introduced in Table \ref{tab:notations}. In this paper, we only consider symmetric meta-paths with the length of 2, denoted as $\langle region, factor, region \rangle$. As shown in Figure \ref{fig:example}, the interactions between regions will be modeled by the shared features. For example, region 1 and region 2 are connected as they both belong to high-income area, while region 1 and region 3 will share the information of urbanization as both of them have a large number of urban facilities. Also, region 1, region 4, and region 5 are neighbors to each other, which share the attribute of short distances on the geographic level. {The choice of various meta-paths will be further discussed in Section \ref{subsec9}.}

\section{Model}\label{sec3}

In this section, we introduce the proposed model \textbf{\underline{S}}patial-\textbf{\underline{T}}emporal \textbf{\underline{M}}eta-path guided \textbf{\underline{E}}xplainable \textbf{\underline{C}}rime prediction (\textbf{STMEC}). STMEC is constructed by four major components: temporal information embedding, meta-path instance encoder, similarity based intra-path aggregation and attention based inter-path aggregation. We first provide an overview of STMEC, and the details of each module are elaborated in the following subsections.

\subsection{Overview of STMEC Architecture}\label{subsec1}
\begin{figure*}[t]
  \centering
  \includegraphics[width=\linewidth]{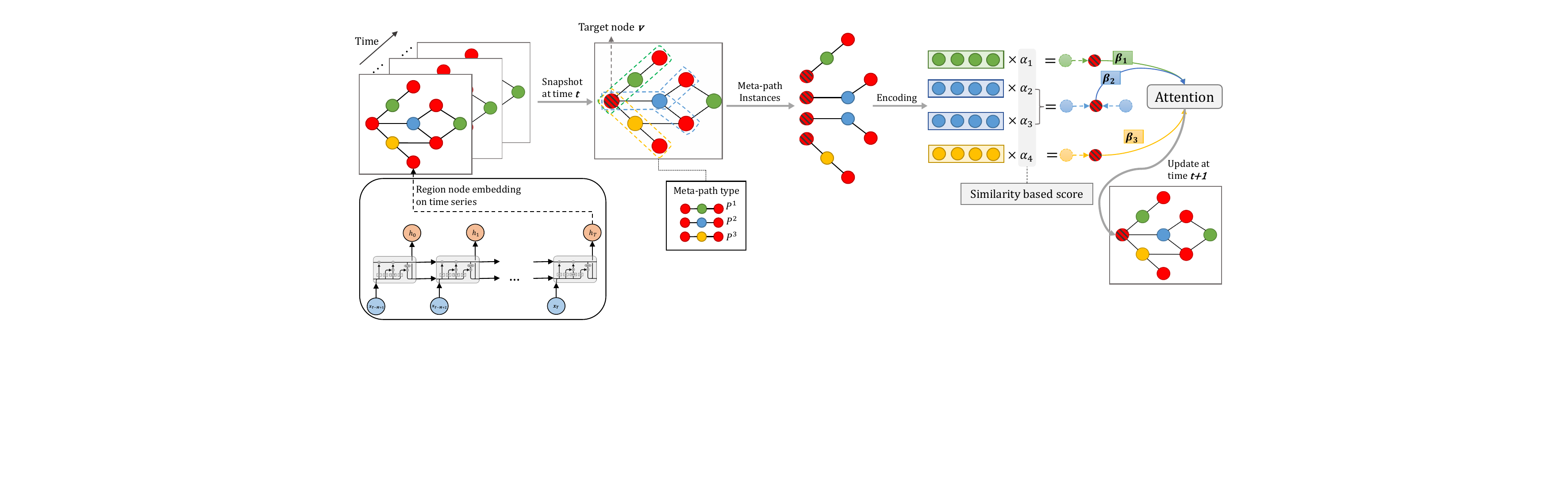}
  \caption{The architecure of Spatial-Temporal Meta-path Guided Explainable Crime Prediction framework. Here is an example based on three types of meta-paths as denoted in Figure \ref{fig:example}. In this example, the red nodes represent regions, while the nodes with other colors are different attributes.}
  \label{fig:framework}
\end{figure*}

The overall architecture of the proposed STMEC model is shown in Figure \ref{fig:framework}. Different from existing graph based crime prediction models which represent the links between regions by either geographical distance or similarity of points of interest (POI), we model the interaction between regions via an intermediate node of a certain attribute type (e.g., demographic distribution). These interactions are characterized as symmetric meta-paths such as $\langle region, factor, region \rangle$ on a meta-path based graph. Given the temporal properties of this meta-path based graph, the embedding of each region is learned and updated along the time steps.
To achieve better representation learning for regions, the procedure is decomposed into four steps.
Firstly, the initial embedding of each region is obtained by learning the latent representations of local crime trends with a recurrent neural network (i.e., Long Short-Term Memory networks (LSTM)).
Secondly, for each meta-path instance, we integrate the embeddings of attributes and regions along the path.
Thirdly, similarity based weights are computed and assigned to meta-path instances to improve the interpretability and accuracy of the model. In this step, the weights are conditioned on the attribute distributions between the pair of regions, and we obtain the representation via weighted aggregation of all meta-path instances of the same type.
Finally, the final representation of aggregated meta-paths is generated via an attention mechanism to facilitate predictions. The attention weights help reason out the contributions of different attribute types, providing insights into crime prevention.

\subsection{Temporal Information Embedding}\label{subsec2}
We firstly learn the latent representation of each region at every time step from the criminal records. For sequential data modeling, LSTM is known for its ability to capture both short-term and long-term dependencies when compared with other RNN based models \cite{Felix01}. Thus, in this task, we deploy a time-series oriented LSTM to learn latent representations from the given sequence of crime records.

For each region $r_i$, given a list of crime records containing $T$ daily data points $\mathcal{Y}_{i}=\{\mathbf{y}_{i}^{1}, \mathbf{y}_{i}^{2}, \ldots, \mathbf{y}_{i}^{t}, \ldots \mathbf{y}_{i}^{T}\} \in \mathbb{R}^{T \times C}$, where $\mathbf{y}_{i}^{t} \in \mathbb{R}^{C}$ is the number of crime records for all crime types $C$ at time step $t$. We aim to initialize the representation of region $r_i$ at time step $t$ based on the crime records from previous $M$ time steps, which is denoted as $\mathcal{Y}_{i}^{M}=\{\mathbf{y}_{i}^{t-M+1}, \ldots, \mathbf{y}_{i}^{t}\} \in \mathbb{R}^{M \times C}$. 
By taking a region's features $\mathbf{y}$ at different time steps as the input, the hidden states $\mathbf{h}_{r_i}^{t}$ derived from the LSTM will be utilized as the initial representation of region $r_i$ at time step $t$. An LSTM unit is composed of a forget gate, an input gate, and an output gate, which regulate the flow of information. Formally, the LSTM layer performs the following functions to update the hidden states $\mathbf{h}_{r_i}^{t}$:
\begin{equation}
\label{eq1}
\begin{split}
\mathbf{f}_{r_i}^{t} &= \sigma\left(\mathbf{W}_{xf} \mathbf{y}_{i}^{t}+\mathbf{W}_{hf}\mathbf{h}_{r_i}^{t-1}+\mathbf{b}_{f}\right)\\
\mathbf{i}_{r_i}^{t} &= \sigma\left(\mathbf{W}_{xi} \mathbf{y}_{i}^{t}+\mathbf{W}_{hi}\mathbf{h}_{r_i}^{t-1}+\mathbf{b}_{i}\right)\\
\tilde{\mathbf{c}}_{r_i}^{t} &= \tanh{\left(\mathbf{W}_{xc}\mathbf{y}_{i}^{t}+\mathbf{W}_{hc}\mathbf{h}_{r_i}^{t-1}+\mathbf{b}_{c}\right)}\\
\mathbf{o}_{r_i}^{t} &= \sigma\left(\mathbf{W}_{xo}\mathbf{y}_{i}^{t}+\mathbf{W}_{ho}\mathbf{h}_{r_i}^{t-1}+\mathbf{b}_{o}\right)\\
\mathbf{c}_{r_i}^{t} &= \mathbf{f}_{r_i}^{t} \odot \mathbf{c}_{r_i}^{t-1}+\mathbf{i}_{r_i}^{t} \odot \tilde{\mathbf{c}}_{r_i}^{t}\\
\mathbf{h}_{r_i}^{t} &= \mathbf{o}_{r_i}^{t} \odot \tanh{\left(\mathbf{c}_{r_i}^{t}\right)}
\end{split}
\end{equation}
where $\mathbf{y}_{i}^{t}$ is the input at time $t$ with dimension $d_y$, and $\mathbf{h}_{r_i}^{t}$ is the hidden state at time $t$ with dimension $d_s$. Furthermore, the input gate, forget gate, output gate and cell state are denoted as $\mathbf{i}_{r_i}^{t}$, $\mathbf{f}_{r_i}^{t}$, $\mathbf{o}_{r_i}^{t}$, and $\mathbf{c}_{r_i}^{t}$, respectively. Each of the $\mathbf{W}_{x*} \in \mathbb{R}^{{d_y} \times {d_s}}$ is the weight matrix to control information flow from input to LSTM cell, while each of the $\mathbf{W}_{h*} \in \mathbb{R}^{{d_s} \times {d_s}}$ is the weight matrix to transform the previous states $\mathbf{h}_{r_i}^{t-1}$ to LSTM cell. Also, $\mathbf{b}_* \in \mathbb{R}^{d_s}$ represents the bias term. The sigmoid function denoted as $\sigma$ helps the LSTM cell to update or forget the data. For notation simplicity, we denote Eq. \ref{eq1} as $\mathbf{h}_{r_i}^{t} = \mathrm{LSTM}(*, \mathbf{c}_{r_i}^{t-1}, \mathbf{h}_{r_i}^{t-1})$ in the following subsections.

\subsection{Meta-path Embedding}\label{subsec3}
In crime prediction tasks, external features depicting the community profiles are normally considered as auxiliary knowledge to complement the machine learning methods. For instance, demographics, income level, and human behavioral factors indicating the socioeconomic characteristics \cite{10.1145/2663204.2663254, Fateha01} are commonly used to improve the crime incident prediction. However, how to leverage heterogeneous and ubiquitous information effectively is always a challenge. In this case, the meta-path based graph is considered to represent heterogeneous information as it can model various relationships between different types of objects intuitively and precisely. Different meta-path based embedding techniques, such as the random-walk based methods \cite{10.1145/3097983.3098036, 10.1109/TKDE.2018.2833443} (i.e., Metapath2vec and HERec), are proposed to aggregate information from neighbors along the paths but are generally more focused on the structural information of each meta-path type and simply treat the inherent node attributes as features. However, in the crime prediction context, how the regions correlate and interact with those features are crucial to accurate predictions. In general, criminal activities always interact with different socioeconomic factors, and the relations can be modeled in a symmetric structure denoted as $\langle region, factor, region\rangle$. This structure protects the diversity of auxiliary information and maps the correlation between regions in terms of different views. Thus, inspired by \cite{10.1145/3366423.3380297}, we distill comprehensive semantics by differentiating meta-path instances into different categories and integrating information by meta-path aggregation methods. In this work, we consider six factors as potential causes of crime-prone communities, including demographics, income level, job type, journey to work, urban facilities, and geographical distance. Each factor is represented by a vector that describes the distribution of its attributes. For example, a feature vector for urban facilities carries the proportion of recreation, residential, and public safety facilities. 

{It is also worth mentioning that, we consider only symmetric meta-paths for capturing region-wise similarity. For example, if two regions have similar income statistics, they will be connected by a virtual meta-path link as $\langle region, income, region \rangle$ (RIR), and it can be extended to RIRIR which has a length of more than 2, but it will reduce the training efficiency as more steps of aggregation are involved. Also, as we focus on learning the importance of different factors to the prediction results, asymmetric meta-path like $\langle region, income, region, job, region \rangle$ (RIRJR) is likely to introduce more noise in the meta-path instance, and will bring difficulties in quantifying the contributions of different socioeconomic factors like income and job in this case. Furthermore, asymmetric meta-paths come with a higher demand on the feature engineering process that heavily depends on domain expertise.} 

\subsection{Meta-path Instance Aggregation}\label{subsec4}
Given target region $r_i$ for which we want to predict the crime incidents in the next time step $t+1$, we first take the snapshot of the entire meta-path based graph at time $t$. In the snapshot, we can find a candidate region $r_j$ connects to $r_i$ via an intermediate node indicating a certain factor. As a meta-path is defined by $\langle region, factor, region\rangle$, we denote the meta-path instance as $p(r_i,r_j)$ and the intermediate node as $m^{p(r_i,r_j)}$, where the type of meta-path instance $P_A$ is determined by the type of intermediate node $A \in \mathcal{A}$. For each type of meta-path, the basic idea is to find all meta-path instances containing target node $r_i$ and measure the priority of each instance. This is achieved by quantifying the similarity between the target region and another region w.r.t. their relationships towards the intermediate node. For instance, if region $r_i$ and region $r_j$ both belong to low-income areas, they will be connected by the intermediate node $income$. If $r_j$ has similar measurements regarding the income-related features (i.e., mean, median, and variance in our case), we would consider $r_j$ as an important candidate for predicting events that happen in $r_i$ when modeling the $\langle region, income, region\rangle$ meta-path. Section 3.4 further introduces the computation of pairwise similarities between two regions in a meta-path instance.

We denote region $r_i$'s feature vector of a certain factor as $\mathbf{m}_{r_i}$. {For example, the feature vector that describes the region's income level comprises of the median, mean, variance, and stand deviation of the local income.} To embed the information along a particular path instance into a low-dimensional vector, we concatenate the node features to preserve the heterogeneity \cite{yili01, 8731426}. Following the concatenation, we apply a linear transformation to map the sequence of features into the same latent space. For a meta-path instance $p(r_i,r_j)$, {we denote its embedding at time $t$ as follows}:
\begin{equation}
\label{eq2}
\mathbf{h}_{p(r_i,r_j)}=\mathbf{W}_p \cdot \mathrm{CONCAT}\left(\mathbf{h}_{r_i}^t,\mathbf{m}_{r_i}^{t},\mathbf{m}_{r_j}^{t},\mathbf{h}_{r_j}^t\right)
\end{equation}
where $\mathbf{h}_{p(r_i,r_j)} \in \mathbb{R}^{d_e}$ is the meta-path instance embedding, and $\mathbf{W}_p \in \mathbb{R}^{d_e \times d'}$ is the learnable parameter. Here, $d'$ is the dimension of the concatenated feature vector and $d_e$ is the embedding size of the concatenated features. In addition,
$\mathbf{h}_{r_i}^t$ is the temporal embedding of region $r_i$ learnt from the previous $M$ time steps.

To aggregate information from meta-path instances for target node $r_i$, we can perform a weighted sum of the instances for each type of meta-path $P_A$:
\begin{equation}
\label{eq3}
\mathbf{h}_{r_i}^{P_A}= \sum_{p(r_i,r_*) \in {P_A}} s(r_i,r_*) \cdot \mathbf{h}_{p(r_i,*)}
\end{equation}
where $r_*$ denotes the candidate region connected by a meta-path instance belonging to type $P_A$ and ends at region $r_i$. The similarity score $s(r_i,r_*)$ is computed by our proposed distribution-aware PathSim. In what follows, we present the innovative design of this path based similarity metric.

\subsection{Distribution-aware PathSim}\label{subsec5}
PathSim \cite{10.14778/3402707.3402736} is a well-established method that outstands for its ability to capture the subtle semantic similarities between objects in symmetric meta-paths. The original PathSim is presented below:
\begin{equation}
\label{ps}
\textcolor{black}{s'\left(r_i,r_j\right) = \frac{2 \times \lvert p_{r_i \rightarrow r_j}:p_{r_i \rightarrow r_j} \in \mathcal{P}\rvert}{\lvert p_{r_i \rightarrow r_i}:p_{r_i \rightarrow r_i} \in \mathcal{P}\rvert+\lvert p_{r_j \rightarrow r_j}:p_{r_j \rightarrow r_j} \in \mathcal{P}\rvert}}
\end{equation}
\textcolor{black}{However, the original PathSim relies only on occurrences of categorical node features and lacks an effective approach to take into account the fine-grained, sub-divided categorical features when computing two regions' similarity in the crime prediction context.} For example, in the meta-path with $urban\, facilities$ as the intermediate node, a vector containing proportions of different building facilities has been used to indicate the urbanization level of a region. A building facility is described as a commonplace that accommodates diverse activities, such as residential, civic, educational, or commercial facilities. These types of facilities are further subdivided into more specific facility types like high school and primary school in the category of educational facilities. Such sub-divided categorical features are commonly observed within the ubiquitous data, making the traditional PathSim incompatible. \textcolor{black}{This is because the similarity between regions w.r.t. different fine-grained categorical features are treated evenly in PathSim, which leads to significant information loss.} Hence, we propose a distribution-aware PathSim to complement the similarity metrics. The distribution-aware PathSim $s(r_i, r_j)$ for two regions is given by:
\begin{equation}
\label{eq6}
\begin{gathered}
s\left(r_i,r_j\right) = \sum_{z \in Z}\frac{\lvert z\rvert}{\lvert Z\rvert} \cdot s\left(r_i,z,r_j\right)\\
s\left(r_i,z,r_j\right) = \frac{2 \times \lvert p_{r_i \rightarrow r_j}:p_{r_i \rightarrow r_j} \in \mathcal{P}_z\rvert}{\lvert p_{r_i \rightarrow r_i}:p_{r_i \rightarrow r_i} \in \mathcal{P}_z\rvert+\lvert p_{r_j \rightarrow r_j}:p_{r_j \rightarrow r_j} \in \mathcal{P}_z\rvert}
\end{gathered}
\end{equation}
{where $z$ represents the facility category (e.g., $Educational$), $Z$ indicates all types of urban facilities, and $\mathcal{P}_z$ denotes the meta-path connected by the category $z$. Here, $\lvert z\rvert$ and $\lvert Z\rvert$ denote the number of facility types in category $z$ and the number of facility types across all categories. For example, region $r_1$ in Table \ref{tab:pathsim} has two categories of urban facilities as $Educational$ and $Recreational$, where $Elementary\ School$ and $High\ School$ are the 2 facility types of $Educational$, and there are in total 4 facility types: $Elementary\ School$, $High\ School$, $Zoo$, and $Pool$ across all categories. Thus, $\frac{\lvert z\rvert}{\lvert Z\rvert}$ can represent the proportion of $Educational$ in all facility types as $\frac{2}{4}$.} Also, $s\left(r_i,z,r_j\right)$ represents the similarity score between region $r_i$ and $r_j$ in terms of the category $z$, while $s\left(r_i,r_j\right)$ is the similarity between two regions concerning all categories $Z$. The measurement is mainly based on the number of path instances between two regions, which is denoted as $\lvert p_{r_i \rightarrow r_j}\rvert$.

\begin{table}
\footnotesize
  \centering
  \begin{tabular}{ccccc}
    \toprule
    \multirow{2}{*}{Region} & \multicolumn{2}{c}{Educational} &\multicolumn{2}{c}{Recreational}\\
    \cmidrule(lr){2-3}\cmidrule(lr){4-5}
    & Elementary School & High School & Zoo & Pool\\
    \midrule
    $r_1$ & 2 & 2 & 5 & 5\\
    \midrule
    $r_2$ & 2 & 2 & 0 & 10\\
    \midrule
    $r_3$ & 0 & 0 & 5 & 5\\
    \bottomrule
\end{tabular}
\caption{A toy example to compare the original PathSim and distribution-aware PathSim.}
 \label{tab:pathsim}
\end{table}

To demonstrate how our proposed method works, we use a toy example as shown in Table \ref{tab:pathsim} to illustrate the reliability of the method compared with the original PathSim. The example shows the number of urban facilities in region $r_1$, $r_2$, and $r_3$. The urban facilities can be divided into two categories as $Educational$ and $Recreational$, and each has two facility types as $Elementary\ School$ and $High\ School$, $Zoo$ and $Pool$ respectively. We aim to find the region with the most similar structure as $r_1$. As $r_2$ has both educational and recreational facilities of the same amount, it is more close to the urban structure as $r_1$. PathSim measures the similarity scroes as: $s'(r_1,r_2) = \frac{2 \times \left(2 \times 2 + 2 \times 2 + 5 \times 10\right)}{\left(2 \times 2+2 \times 2+5  \times 5 + 5 \times 5 \right) + \left(2 \times 2+2 \times 2 + 10 \times 10\right)}=0.699$, while $s'(r_1,r_3)=0.926$. The distribution-aware PathSim complements the problem by taking the weighted sum of PathSim in terms of categories. Thus, we compute the similarity score as $s(r_1,r_2) = \frac{2}{4} \cdot 1+\frac{2}{4} \cdot \frac{2\times \left(5 \times 10 \right)}{\left(5 \times 5 + 5 \times 5\right) + \left(10 \times 10\right)}=0.833$, while $s(r_1,r_3)=0.5$. Apparently, our distribution-aware PathSim has stronger consistency with human intuition owing to the consideration of different categories.

\subsection{Attention for Meta-path based Context}\label{subsec6}
Intuitively, different meta-paths carry different semantics in a region-region interaction. In the task of crime prediction, it is hard to identify the leading factor that contributes more to criminal behaviors. Also, for different regions, a meta-path may have varying semantics as it collects information from different instances via the interaction. Hence, we apply a graph attention layer to rank the importance of meta-paths and then summarize the flow of information by weighted sum.

For region $r_i \in I$, we have the summarization of the meta-path $P_A$ denoted as $\mathbf{h}_{r_i}^{P_{A}}$, where $P_{A} \in \mathcal{P}$ and $\mathcal{P}$ is a set of meta-paths containing node $r_i$. First, we transform the meta-path based representations for all nodes $r_i \in I$, and obtain the average value with respect to each meta-path type $P_A$:
\begin{equation}
\label{eq4}
\mathbf{u}^{P_A} = \frac{\sum_{r_i \in I} \tanh\left(\mathbf{M}_u \cdot \mathbf{h}_{r_i}^{P_A}+\mathbf{b}_u\right)}{|I|}
\end{equation}
{where $\mathbf{M}_u \in \mathbb{R}^{d_a \times d_e}$ and $\mathbf{b}_u \in \mathbb{R}^{d_a}$ are learnable parameters.} The number of nodes is denoted by $|I|$.

Then the attention mechanism is utilized to fuse the meta-path based context, the weights are learnt over different types of meta-path as follows:
\begin{equation}
\label{eq5}
\begin{split}
e^{P_A} &= \mathbf{q}^{T} \cdot \mathbf{u}^{P_A}\\
\beta^{P_A} &= \frac{\exp\left(e^{P_A}\right)}{\sum_{P_A \in \mathcal{P}} \exp\left(e^{P_A}\right)}\\
\mathbf{h}_{r_i}^{\mathcal{P}} &= \sum_{P_A \in \mathcal{P}}\beta^{P_A} \cdot \mathbf{h}_{r_i}^{P_A}
\end{split}
\end{equation}
{where $\mathbf{q} \in \mathbb{R}^{d_a}$ is the attention vector to be learned in the training process}, and $\beta^{P_A}$ is the learnt importance score of meta-path $P_A$. The summarized context information based on each type of meta-path is embedded as $\mathbf{h}_{r_i}^{\mathcal{P}}$, which will be projected to a $C$-dimensional output via a dense layer with sigmoid activation. {The final output is the estimated probability of crime $c$ that will happen in region $r_i$, which is denoted as $\hat{y}_{r_i}^{t, c}$}.

\subsection{Training}\label{subsec7}
Our objective is to obtain the value of $\hat{\mathbf{y}}_{r_i}^{t}$ at each time step $t$, where the $c$-th element $\hat{y}_{r_i}^{t,c}$ denotes the probability that crime event of category $c$ will happen in the next time step $t+1$. As this can be viewed as $C$ binary classification tasks, we employ cross entropy as the loss function:
\begin{equation}
\label{eq7}
\mathcal{L}=-\sum_{(r_i, c, t) \in S} y_{r_i}^{t+1, c}\log \hat{y}_{r_i}^{t, c}+\left(1-y_{r_i}^{t+1, c}\right)\log\left(1-\hat{y}_{r_i}^{t, c}\right)
\end{equation}
where $\hat{y}_{r_i}^{t, c}$ is the estimated probability of crime $c$ that will happen in region $r_i$ in the next time slot $t+1$ and $y_{r_i}^{t+1, c}$ is the corresponding ground-truth record at time slot $t+1$. Also, $S$ is the crime event set in the training process. In this work, the model parameters are learnt by minimizing the loss function with Adaptive Moment Estimation (Adam) \cite{Diederik01}.

\section{Experiment}\label{sec4}
\subsection{Experiment Settings}\label{subsec8}
\subsubsection{Datasets}\label{subsubsec1}

We integrate multiple public datasets from various resources in New York City (NYC). The performance of our framework is evaluated via the datasets collected from both 2014 and 2015.

\begin{figure}[ht]
\vspace{-0.3cm}
\centering
    \begin{subfigure}[b]{0.43\textwidth}
         \centering
         \includegraphics[width=\textwidth]{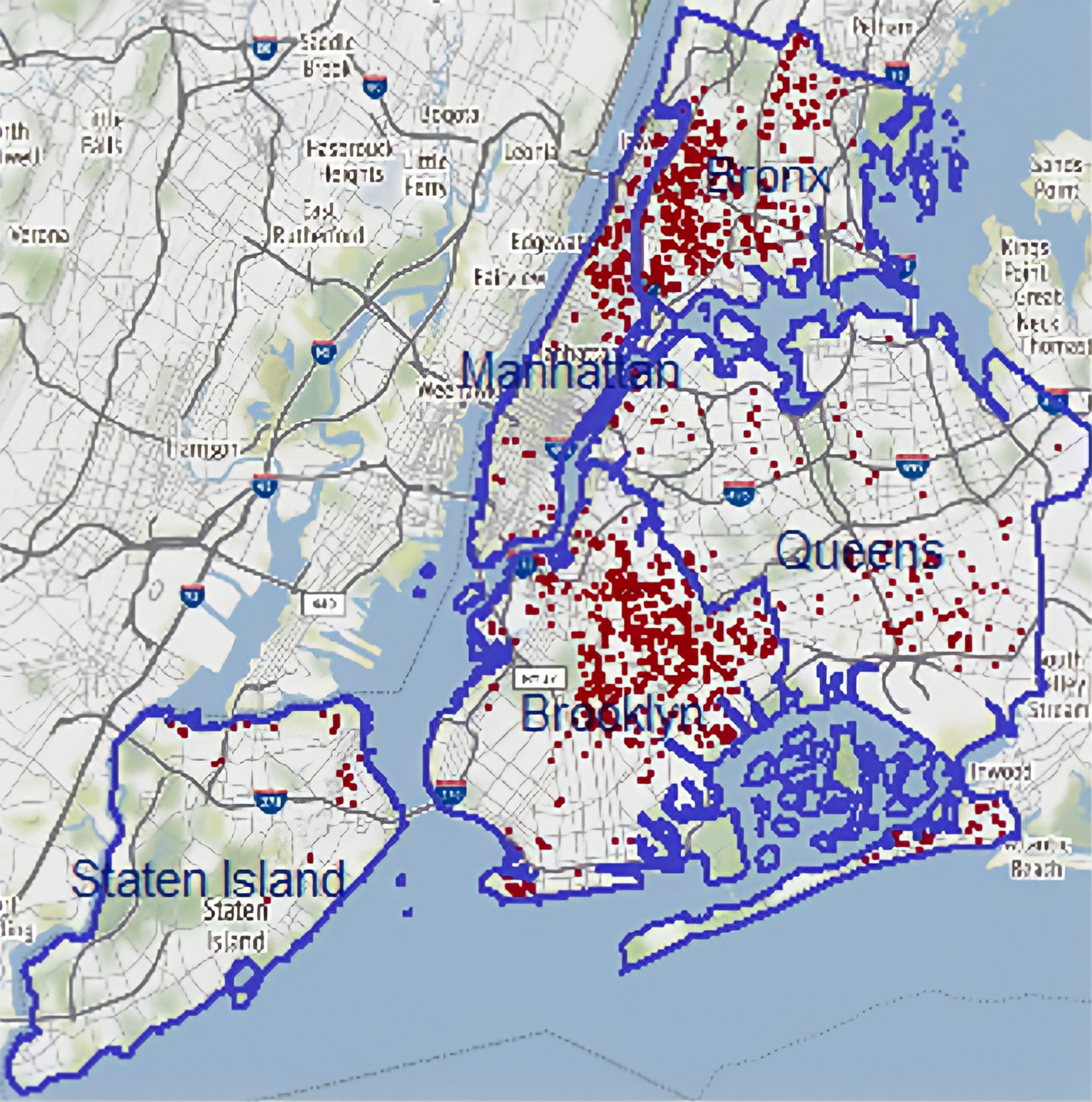}
         \caption{Gun Violence 2014}
         \label{fig:gun2014}
     \end{subfigure}
     \hfill
     \begin{subfigure}[b]{0.435\textwidth}
         \centering
         \includegraphics[width=\textwidth]{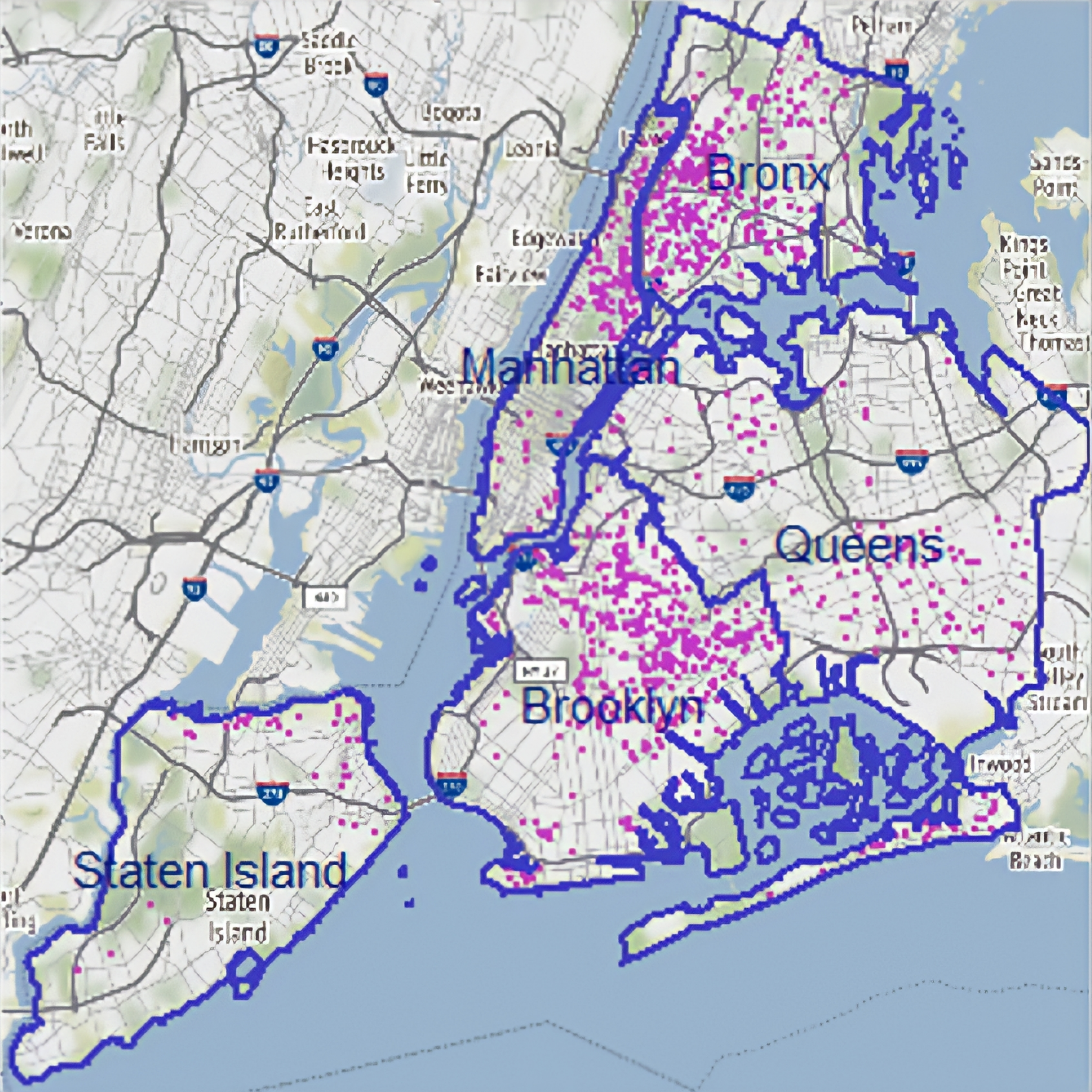}
         \caption{Gun Violence 2015}
         \label{fig:gun2015}
     \end{subfigure}
    \caption{Gun Violence Distribution in NYC.}
    \label{fig:gun crime}
\end{figure}

(1) \textbf{Census Data} \cite{MuonNeutrino17}: There are 2161 records of census tracts in the 2015 American Community Survey containing information about population, commuting ways, job type, and income level.

(2) \textbf{Points of Interest (POI)} \cite{datapoi}: We collect more than 15,000 records of commonplaces, which includes 13 facility domains like Residential Places, Education Facility, Cultural Facility, Recreational Facility, Social Services Facilities. After allocating the number of data points to the corresponding region by each type, the distribution of points in each region can reflect the level of urbanization.

(3) \textbf{Crime Data} \cite{datacrime}: We evaluate the STMEC model on the crime data collected from Jan 1, 2014, to Dec 31, 2015 (2*365 days) in NYC. For the prediction tasks, we select the top 10 most common crimes. Apart from this, we also investigate the dangerous weapon related crimes, as more than 200 people in the United States are wounded or killed by gun violence every day \cite{10.1001/jamainternmed.2016.8245}. As shown in Figure \ref{fig:gun crime}, gun violence in NYC usually happens in Brooklyn and the borders of Manhattan and Bronx, which indicates that gun violence is highly related to geographical locations.

There are in total 71 districts in NYC based on the partition rule that is valid from 2013 to 2020, and our experiments are conducted on the district level. Hence, the ubiquitous data has been integrated into each district to represent a higher level of various characteristics. The statistics of crime data and POI are shown in Table \ref{tab:crimedata} and Table \ref{tab:poi}.

\begin{table}[ht]
\vspace{-0.3cm}
\renewcommand{\arraystretch}{1.3}
\centering
\footnotesize
  \begin{tabular}{M{0.13\textwidth}M{0.21\textwidth}M{0.1\textwidth}M{0.1\textwidth}M{0.1\textwidth}M{0.1\textwidth}}
    \toprule
    \multirow{2}{0.13\textwidth}[-0.5em]{\centering \textbf{Data Source}} & \multirow{2}{0.14\textwidth}[-0.5em]{\centering \textbf{Category}} & \multicolumn{2}{M{0.2\textwidth}}{\textbf{Avg. Days with Crimes}} & \multicolumn{2}{M{0.2\textwidth}}{\textbf{Avg. Days without Crimes}}\\
    \cmidrule(lr){3-4}\cmidrule(lr){5-6}
    & &  \textbf{2014} & \textbf{2015} & \textbf{2014} & \textbf{2015}\\
    \midrule
    \multirow{11}{0.13\textwidth}[-2em]{\centering \textbf{NYC Crime Reports 2014 and 2015}} & Petite Larceny & 296 & 294 & 69 & 71\\
    \cmidrule(l){2-6}
    & Harassment & 281 & 281 & 84 & 84 \\
    \cmidrule(l){2-6}
    & Assault & 258 & 258 & 107 & 107 \\
    \cmidrule(l){2-6}
    & Criminal Mischief & 254 & 256 & 111 & 109\\
    \cmidrule(l){2-6}
    & Grand Larceny & 250 & 249 & 115 & 116 \\
    \cmidrule(l){2-6}
    & Dangerous Drugs & 167 & 154 & 198 & 211 \\
    \cmidrule(l){2-6}
    & Against Public Order & 173 & 182 & 192 & 183\\
    \cmidrule(l){2-6}
    & Felony Assault& 165 & 165 & 200 & 200 \\
    \cmidrule(l){2-6}
    & Robbery & 152 & 155 & 213 & 210\\
    \cmidrule(l){2-6}
    & Burglary & 155 & 146 & 210 & 219\\
    \cmidrule(l){2-6}
    & Dangerous Weapons & 106 & 105 & 259 & 260\\
    \bottomrule
\end{tabular}
\caption{Data statistics of crime records: average days with and without crimes across all regions.}
 \label{tab:crimedata}
 \vspace{-0.5cm}
\end{table}

\begin{table}[ht]
\vspace{-0.3cm}
\renewcommand{\arraystretch}{1.3}
\centering
\footnotesize
  \begin{tabular}{M{0.12\textwidth}M{0.12\textwidth}M{0.08\textwidth}M{0.08\textwidth}M{0.13\textwidth}M{0.08\textwidth}M{0.08\textwidth}}
    \toprule
    \textbf{Data Source} & \textbf{Category} & \textbf{\#(2014)} & \textbf{\#(2015)} & \textbf{Category} & \textbf{\#(2014)} & \textbf{\#(2015)}\\
    \midrule
    \multirow{6}{0.12\textwidth}[-3em]{\centering \textbf{NYC POI 2014 and 2015}} & Commercial & 711 & 724 & Cultural & 520 & 522\\
    \cmidrule(l){2-7}
    & Education & 3037 & 3472 & Government & 662 & 723\\
    \cmidrule(l){2-7}
    & Health Services & 215 & 220 & Miscellaneous & 648 & 651\\
    \cmidrule(l){2-7}
    & Public Safety & 592 & 596 & Recreational & 2458 & 2481\\
    \cmidrule(l){2-7}
    & Religious & 1164 & 1201 & Residential & 2930 & 2967\\
    \cmidrule(l){2-7}
    & Social Services	& 1395 & 1409 & Transportation & 455 & 542\\
   \cmidrule(l){2-7}
    & Water & 281 & 281\\
    \bottomrule
\end{tabular}
\caption{Data statistics of POI: number of POI of different categories in NYC 2014 and 2015.}
 \label{tab:poi}
 \vspace{-0.5cm}
\end{table}

\begin{figure}[ht]
\vspace{-0.5cm}
\centering
    \begin{subfigure}[b]{0.47\textwidth}
         \centering
         \includegraphics[width=\textwidth]{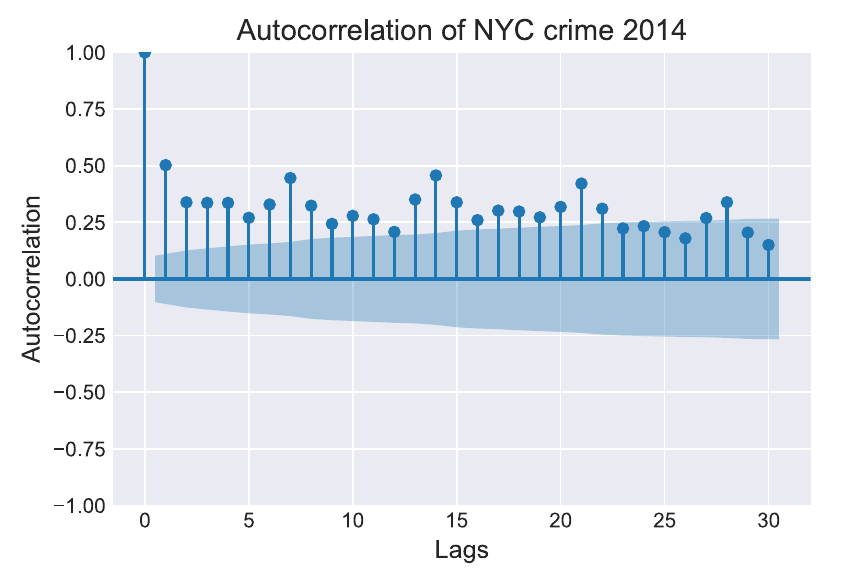}
         \label{fig:acf2014}
     \end{subfigure}
     \hfill
     \begin{subfigure}[b]{0.47\textwidth}
         \centering
         \includegraphics[width=\textwidth]{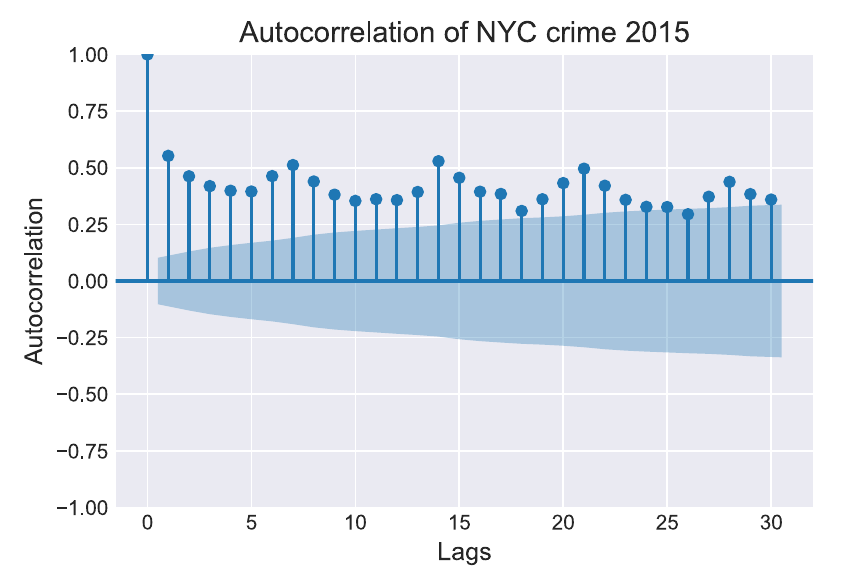}
         \label{fig:acf2015}
     \end{subfigure}
    \vspace{-0.5cm}
    \caption{ACF plot of crime trends in 2014 and 2015.}
    \label{fig:acf}
    \vspace{-0.2cm}
\end{figure}

\begin{figure}[ht]
\vspace{-0.5cm}
\centering
    \begin{subfigure}[b]{0.47\textwidth}
         \centering
         \includegraphics[width=\textwidth]{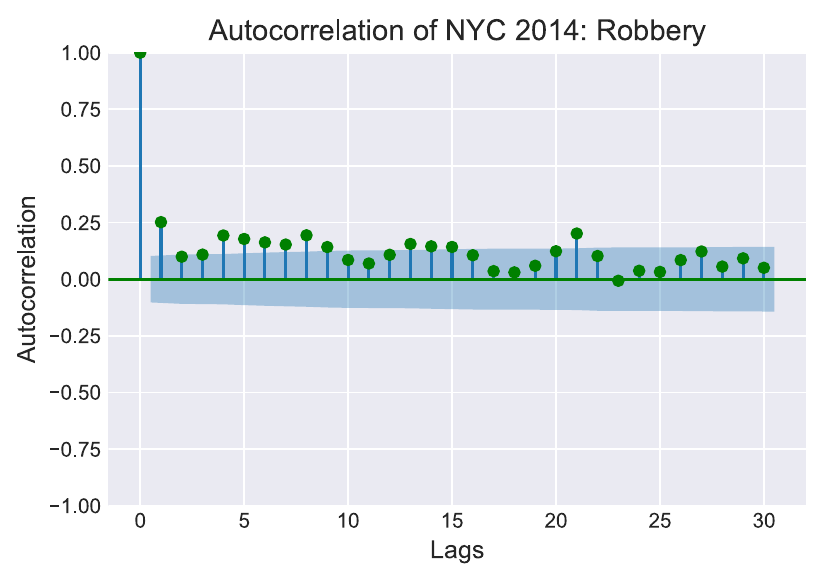}
         \label{fig:acf2014}
     \end{subfigure}
     \hfill
     \begin{subfigure}[b]{0.47\textwidth}
         \centering
         \includegraphics[width=\textwidth]{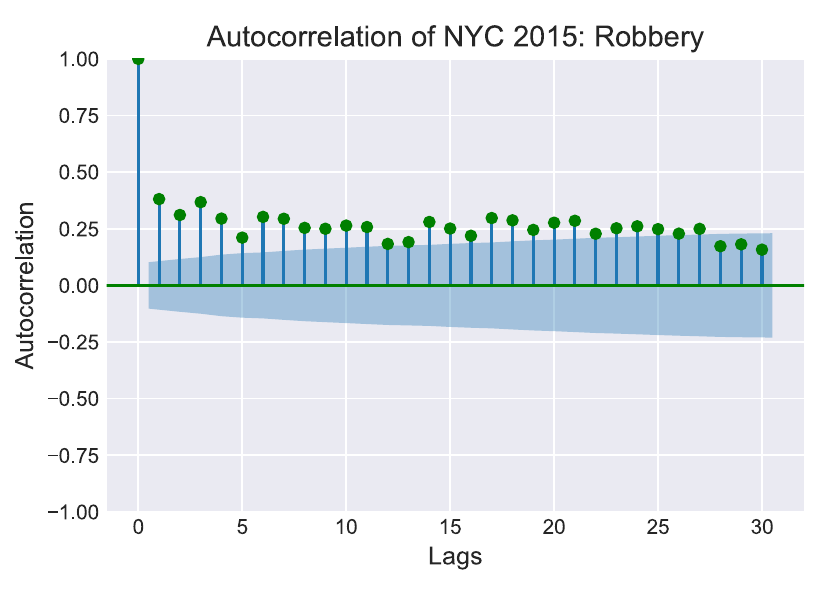}
         \label{fig:acf2015}
     \end{subfigure}
     \vspace{-0.5cm}
    \caption{ACF plot of robbery trends in 2014 and 2015.}
    \label{fig:acf_rb}
    \vspace{-0.7cm}
\end{figure}

\subsubsection{Analysis on Crime Trends}\label{sec:acf}
\textcolor{black}{To better understand the crime trends, we explore the correlation between the crime events and their own lags by using the Autocorrelation Function (ACF). Fig.\ref{fig:acf} shows the non-stationary characteristics of both crime trends, given the slowly decreasing autocorrelation values along the lags. Additionally, the plot also indicates different temporal dependencies on historical crimes across two datasets, as the autocorrelations of different datasets show different levels of sensitivity to their lags. Specifically, compared with crime trends in 2014, crime trends in 2015 show a stronger correlation (i.e., a larger autocorrelation value of recent lags) between the current and recent records.} 

\textcolor{black}{Furthermore, the difference of temporal dependencies among two datasets is also observed from fine-grained crime types. 
For example, in Fig.\ref{fig:acf_rb}, robbery trends of NYC 2014 spikes at lag 21, indicating a strong correlation between the current crime events and the one happened 21 days before, while a weak correlation is observed between the current crime events and recent lags (lag 2 and lag 3). However, robbery trends of 2015 show a stronger temporal dependency across 4 weeks' lags. As we will further discuss in Section \ref{time_w}, these findings suggest different model sensitivity to varying time window sizes in different datasets.}

\subsubsection{Evaluations}\label{subsubsec2}

As an evaluation of the stability and reliability of model performance for different datasets, the experiments are done for crime events in 2014 and 2015 respectively. To ensure the consistency of training and test for STMEC and all baseline models, all prediction tasks start from 29\textsuperscript{th} Jan of each year, and {the data is split into 75\% (255 days) for training, 5\% (18 days) for validation, and 20\% (64 days) for test.} The choice of the commencement date is based on the optional time window used for crime prediction tasks. To compare STMEC with state-of-the-art baselines, we use four types of evaluation metrics to fully investigate the performance of each model on the test set:
\begin{itemize}
\item (1) \textbf{Macro-F1} and \textbf{Micro-F1 score}: To measure the performance of the model across different categories (i.e., Robbery and Grand Larceny) of crime events, we utilize Macro-F1 and Micro-F1 \cite{10.1145/3292500.3330790}. Models with better overall performance usually have higher Micro-F1 and Macro-F1 scores.
\item (2) \textbf{Macro-Recall} and \textbf{Micro-Recall score}: Considering that criminal activities can lead to an immeasurable loss to the society, the ideal predictor is supposed to minimize the number of false negatives, that is, we want to correctly identify as many risky activities as possible. Thus, the score of recall is a crucial metric in crime prediction. Similar to Macro and Micro F1 score, the Macro and Micro Recall \cite{10.1145/3197026.3197055} is computed by averaging the performance across different categories.
\item (3) \textbf{F1 score}: F1 score is used to measure the prediction accuracy of the model for an individual category of crime, which computes the harmonic mean of precision and recall.
\item (4) \textbf{Recall}: Recall is used to evaluate the performance of the model for a single category of crime events, which suggests the ability to correctly identify the crime events that truly happened.
\end{itemize}

\subsubsection{Baselines}\label{subsubsec3}
The performance of the proposed model will be compared with the state-of-the-art baseline models. Most of the Graph Convolutional Networks (GCN) based models are inherited from the task of traffic prediction, which is also used for predicting crime events in some studies \cite{9050374,9276416,LIANG2021}. Thus, we will compare the performance of our model with these outstanding models, given the purpose of crime prediction is also to capture the spatial and temporal dependence. Please note that we use the same input data including geographic, demographic, and urban-related information for all baseline models but with different data representations concerning different architectures of baseline models.

We briefly introduce the baseline methods for comparison below:
\begin{itemize}
\item \textbf{Multilayer Perceptron (MLP)}: This is the conventional neural network with 3 dense layers to learn the representations from given features.
\item \textbf{GCN} \cite{Thomasn01}: This neural network learns the features by aggregating information from neighboring nodes. To test the performance of this model, the input data has been organized into a graph structure where the nodes represent each region, while the edges are given by the distance between the adjacent regions.
\item \textbf{TGCN} \cite{Ling2020}: This model is normally employed to capture the spatial and temporal dependence simultaneously. It is a joint framework with the GCN and gated recurrent unit (GRU).
\item \textbf{LSTM-GCN} \cite{10.1145/3292500.3330919}: This model is quite similar to TGCN. Instead of using GRU to capture the temporal pattern, this model replaces the component with LSTM unit.
\item \textbf{STGCN} \cite{bing01}: This model is a typical spatial-temporal model but with complete convolutional structures. It can also capture comprehensive spatial-temporal dependence.
\item \textbf{MiST} \cite{10.1145/3308558.3313730}: MiST is a attention based recurrent framework for crime prediction. It surpasses traditional regression based models (i.e., SVR and ARIMA) as well as several state-of-the-art deep learning models. Thus, MiST is a strong baseline to evaluate and performance of our proposed model.
{
\item \textbf{HAGEN} \cite{wang2022hagen}: HAGEN is the most recent graph neural network specifically designed for crime forecasts. This framework captures the crime correlation between regions via distance based and context based (e.g., POI) similarities.
\item \textbf{GSNet} \cite{wang2021gsnet}: GSNet is a homogeneous graph based framework which models different kinds of spatial correlations by different types of context features (e.g., POI similarity, geographical similarity). Based on our problem setting, we have deployed this framework by constructing six homogeneous graphs which correspond to the six factors explored in this paper.}
\end{itemize}

\subsection{Parameter Settings}\label{subsec9}
In our work, the hyperparameters of each baseline model are carefully tuned to ensure optimal results. We implement STMEC with Pytorch architecture. We set the dimension of temporal embedding $d_s$ as $128$ and the number of LSTM layers as $2$. The optimal time window $M$, from which the temporal dynamics are learned, is set as $28$ days for 2015 dataset and $21$ days for 2014 dataset. Also, The embedding size of meta-path instances $d_e$ is set as $64$ and the attention vector size $d_a$ is set as $128$.  Additionally, the learning process is optimized by Adam, with the learning rate 0.0001. {The effect of different hyperparameter setting has been discussed in Section \ref{subsec12}.}

We investigate 6 different types of factors that may affect the criminal activities which are represented as meta-paths: $\langle region, demographic, region\rangle$ (RDR), $\langle region, income, region\rangle$ (RIR), $\langle region, job\: type, region\rangle$ (RJR), $\langle region, commuting\: ways, region\rangle$ (RCR), $\langle region, urbanization, region\rangle$ (RUR), and $\langle region, geographic, region\rangle$ (RGR). The importance of the factors is discussed in the following subsections.

\begin{figure}[H]
\vspace{-0.5cm}
  \centering
  \includegraphics[width=0.9\linewidth]{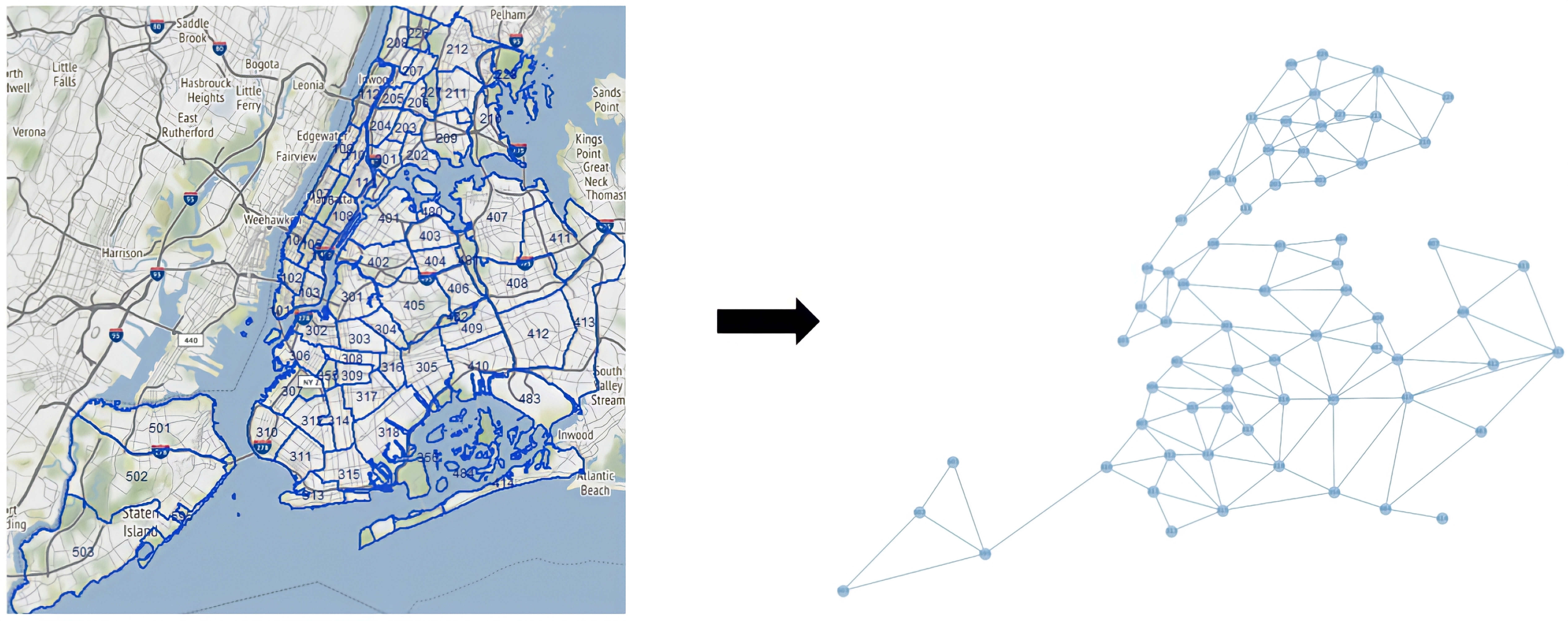}
  \caption{Data representation in GCN based model. Each node represents a region (district) with integrated features (i.e., demographics and urban structures), and the adjacent regions are connected by edges which indicate the distance in between.}
  \label{fig:gcn}
  \vspace{-0.7cm}
\end{figure}

\subsection{Model Performance Comparison}\label{subsec10}
\subsubsection{Overall Performance across All Crime Types}\label{subsubsec4}
As shown in Table \ref{tab:comparison}, the proposed model STMEC outperforms all existing state-of-the-art deep learning models by achieving the best scores across all evaluation metrics. {In the dataset from NYC 2015, we improve 2\%, 6\%, 8.2\%, 5.7\% of the crime prediction framework MiST with respect to Macro-F1, Micro-F1, Macro-Recall, and Micro-Recall scores. Also, our model shows competitive performance even compared to the most recent spatial-temporal frameworks HAGEN and GSNet.} The advantages of our proposed model persist on the other dataset from 2014 as well.  The results prove that our model benefits with better performance by explicitly modeling the interactions between crime events and the possible causes. By comparing the performance of MLP and GCN which ignore the long-term temporal dynamics, no obvious loss of the performance is observed. This result suggests that the crime events in general are largely affected by the most recent trends. Also, the slight improvement of GCN as compared to MLP indicates the importance of geographical factors, as the GCN based model constructs the data as geographic based graphs, which are shown in Figure \ref{fig:gcn}. Consistent with the findings of temporal impacts on crime events, the rest of GRU or LSTM based models which focus more on sequential information cannot significantly improve the performance with comparison to conventional neural networks.

\begin{table*}
\vspace{-0.2cm}
\renewcommand{\arraystretch}{1.2}
\centering
\tiny
   \begin{tabular}{M{0.07\textwidth}M{0.06\textwidth}M{0.035\textwidth}M{0.035\textwidth}M{0.05\textwidth}M{0.05\textwidth}M{0.05\textwidth}M{0.03\textwidth}M{0.05\textwidth}M{0.04\textwidth}M{0.04\textwidth}M{0.01\textwidth}}
    \toprule
    \textbf{Datasets} & \textbf{Metrics} & \textbf{MLP} & \textbf{GCN} & \textbf{TGCN} & \textbf{LSTM-GCN} & \textbf{STGCN} & \textbf{MiST} & {HAGEN} & {GSNet} & \textbf{STMEC}&  \\
    \midrule
    \multirow{4}{*}[-3em]{NYC 2014} & Macro-F1 & 0.794 & 0.802 & 0.803 & 0.792 & 0.803 & 0.797 & \textbf{{0.816}} & {0.813} & \textbf{0.816}&\\
    \cmidrule(l){2-12}
    & Micro-F1 & 0.707 & 0.724 & 0.732 & 0.721 & 0.733 & 0.719 &  {0.761} & {0.759} & \textbf{0.764}&\\
    \cmidrule(l){2-12}
    & Macro-Recall & 0.688 & 0.713 & 0.738 & 0.715 & 0.733 & 0.724 &  {0.786} & {0.786} &\textbf{0.791}&\\
    \cmidrule(l){2-12}
    & Micro-Recall & 0.768 & 0.792 & 0.807 & 0.785 & 0.803 & 0.797 & {0.846} & {0.844} & \textbf{0.849}&\\
    \midrule
    \multirow{4}{*}[-3em]{NYC 2015} & Macro-F1 & 0.793 & 0.798 & 0.784 & 0.793 & 0.795 & 0.796 & {0.813} & {0.810} & \textbf{0.816}&\\
    \cmidrule(l){2-12}
    & Micro-F1 & 0.715 & 0.716 & 0.690 & 0.722 & 0.710 & 0.703 & {0.756} & {0.748} &  \textbf{0.763}&\\
    \cmidrule(l){2-12}
    & Macro-Recall & 0.688 & 0.699 & 0.667 & 0.714 & 0.699 & 0.707 & {0.778} & {0.768} & \textbf{0.789}&\\
    \cmidrule(l){2-12}
    & Micro-Recall & 0.763 & 0.777 & 0.752 & 0.786 & 0.780 & 0.789 & {0.839} & {0.834} & \textbf{0.846}&\\
    \bottomrule
\end{tabular}
\caption{{Performance comparison with baselines.}}
 \label{tab:comparison}
 \vspace{-0.7cm}
\end{table*}

\subsubsection{Performance on Individual Crime Type}\label{subsubsec5}
\begin{figure}[ht]
    \centering
     \begin{subfigure}[b]{0.49\textwidth}
         \centering
         \includegraphics[width=\textwidth]{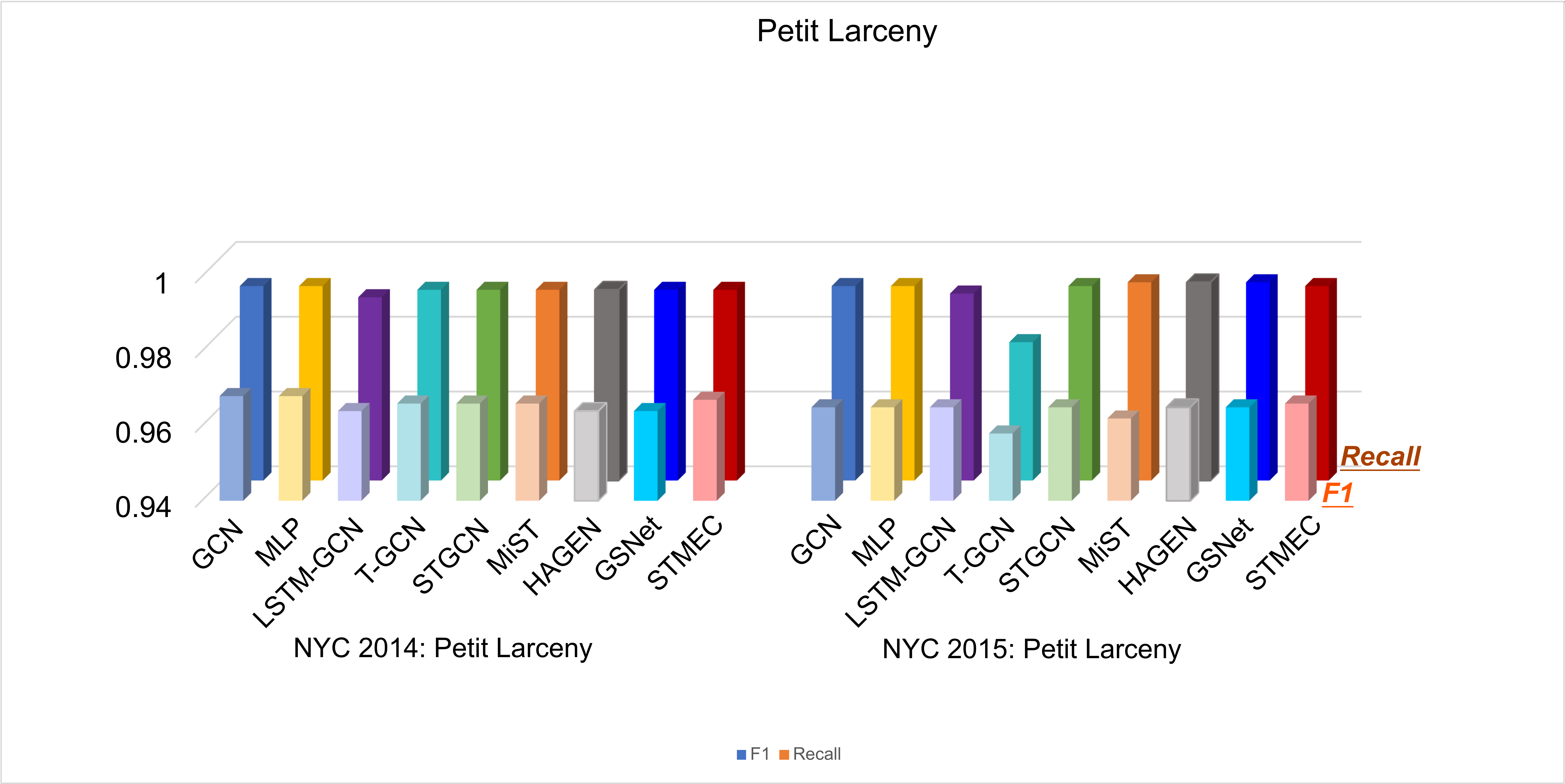}
         \caption{Petit Larceny}
     \end{subfigure}
     \hfill
     \begin{subfigure}[b]{0.49\textwidth}
         \centering
         \includegraphics[width=\textwidth]{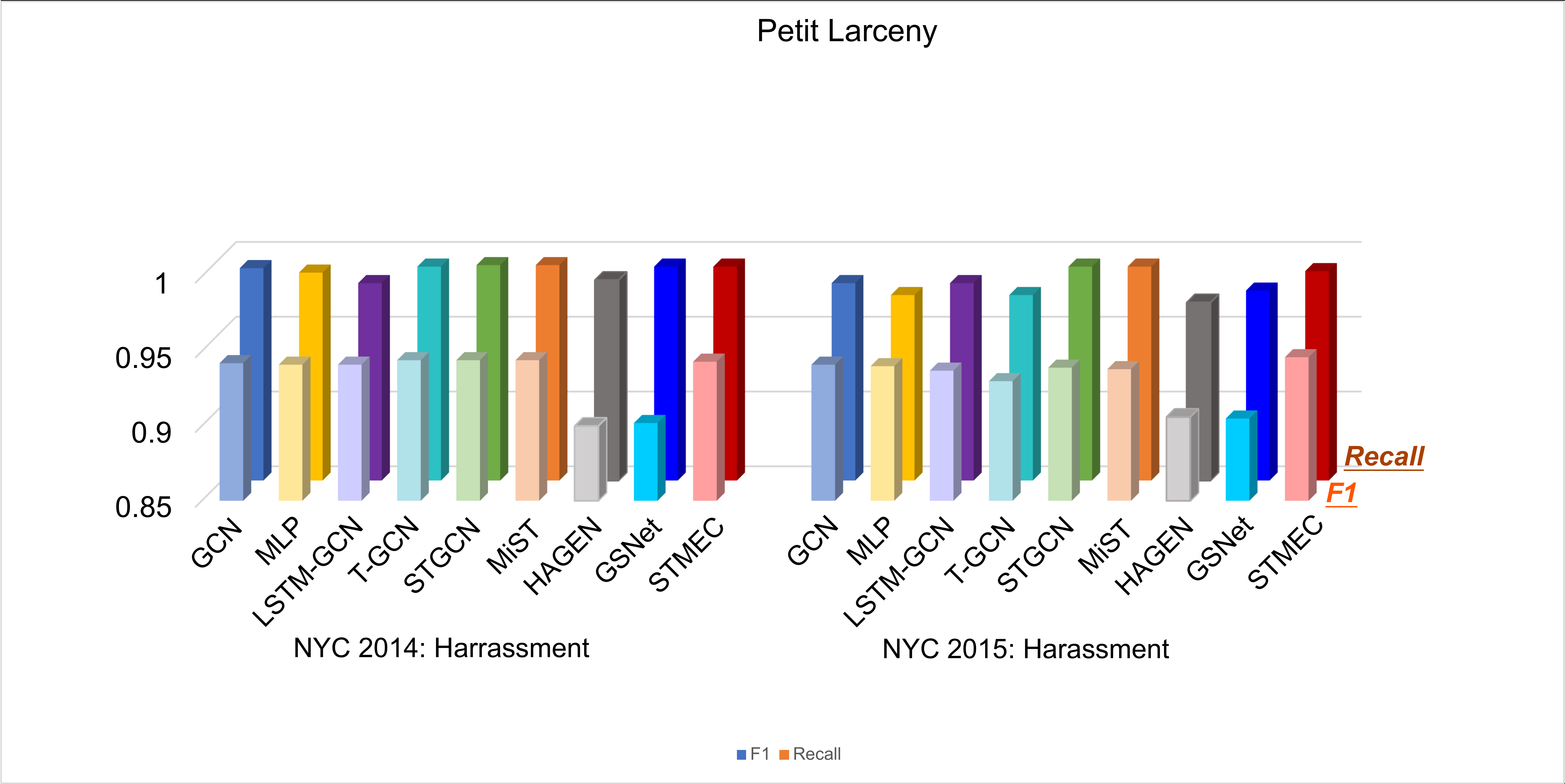}
         \caption{Harassment}
     \end{subfigure}
     \hfill
     \begin{subfigure}[b]{0.49\textwidth}
         \centering
         \includegraphics[width=\textwidth]{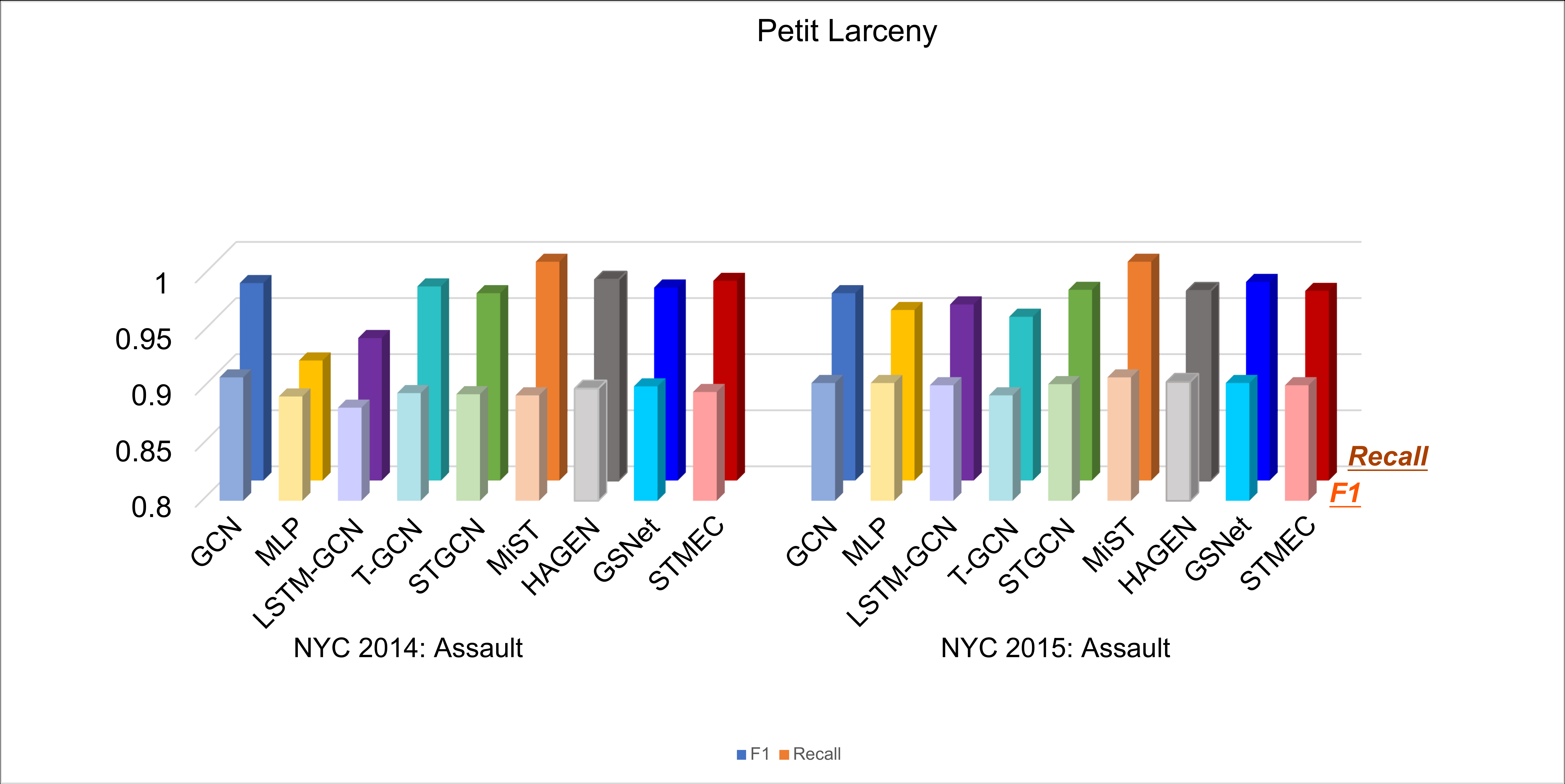}
         \caption{Assault}
     \end{subfigure}
     \hfill
     \begin{subfigure}[b]{0.49\textwidth}
         \centering
         \includegraphics[width=\textwidth]{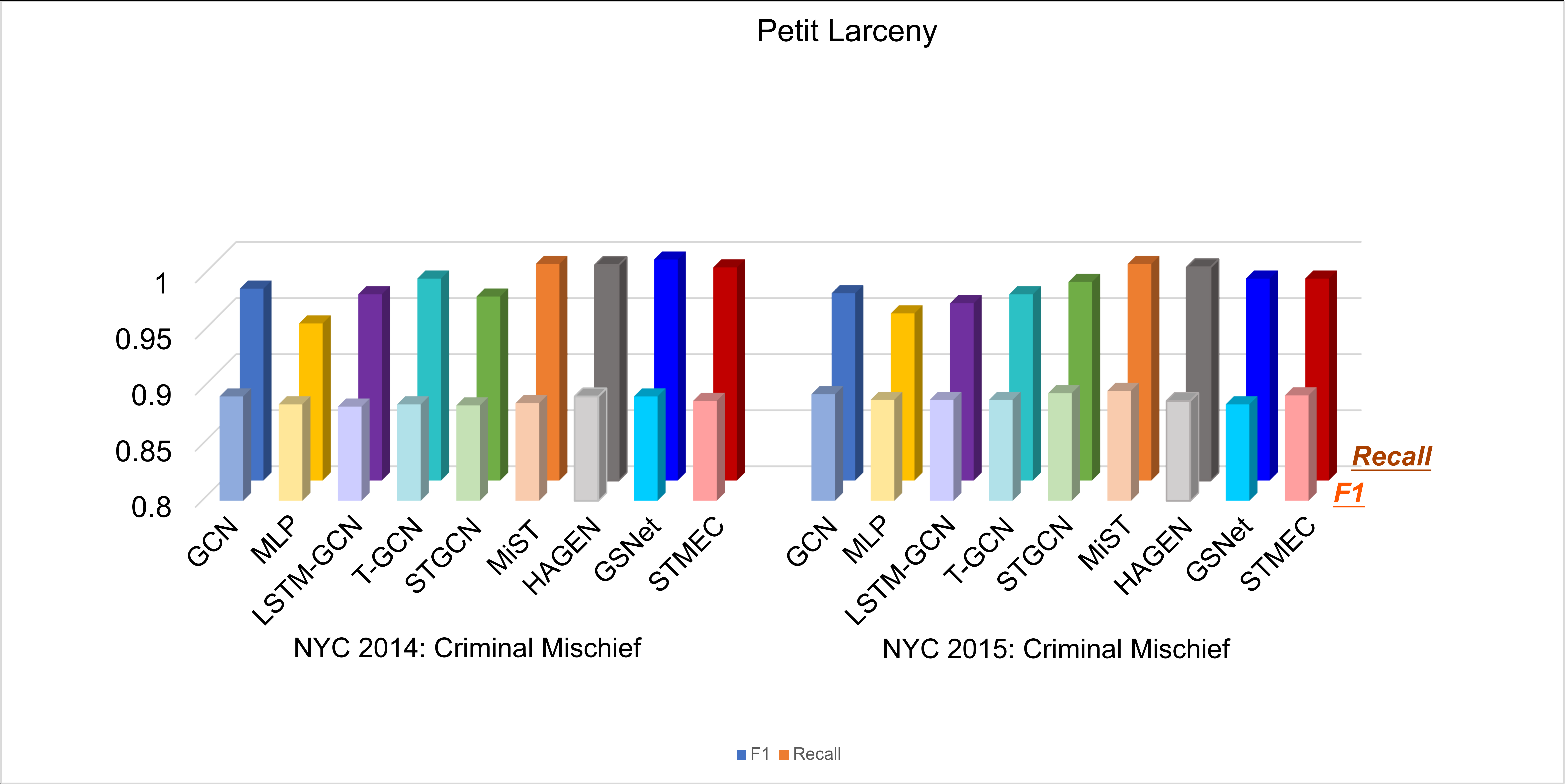}
         \caption{Criminal Mischief}
     \end{subfigure}
     \hfill
     \begin{subfigure}[b]{0.49\textwidth}
         \centering
         \includegraphics[width=\textwidth]{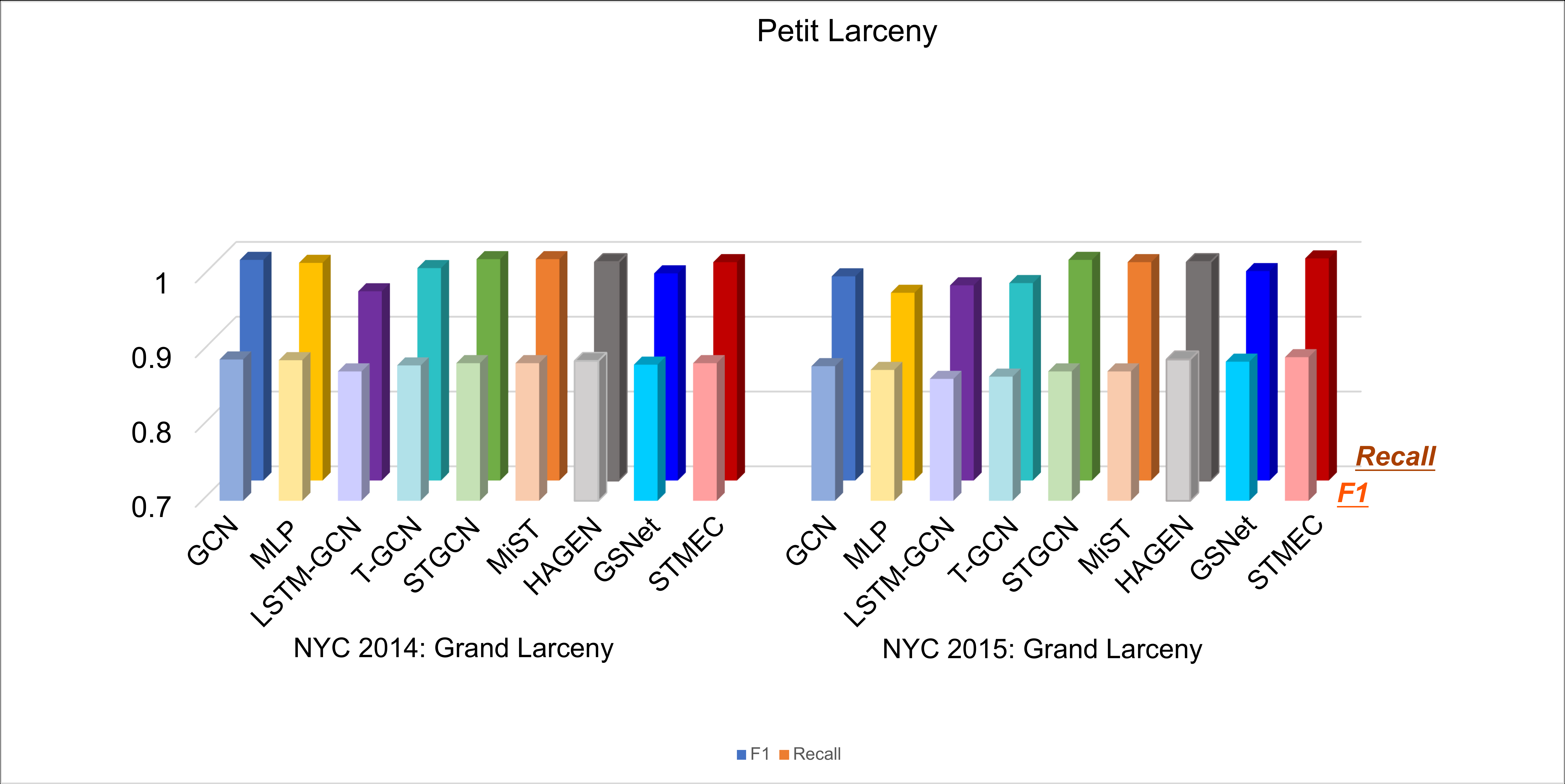}
         \caption{Grand Larceny}
     \end{subfigure}
     \hfill
     \begin{subfigure}[b]{0.49\textwidth}
         \centering
         \includegraphics[width=\textwidth]{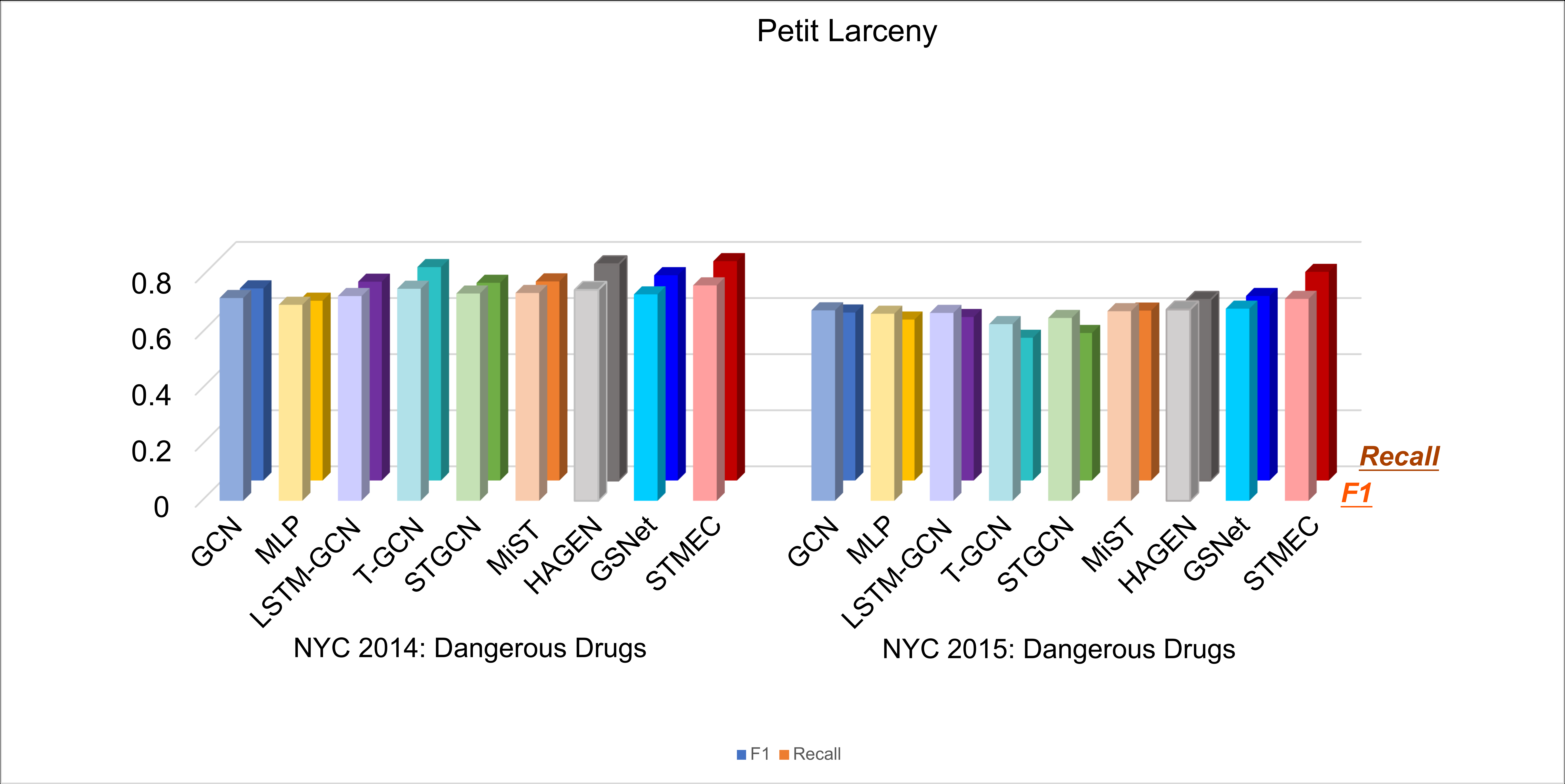}
         \caption{Dangerous Drugs}
     \end{subfigure}
     \begin{subfigure}[b]{0.49\textwidth}
         \centering
         \includegraphics[width=\textwidth]{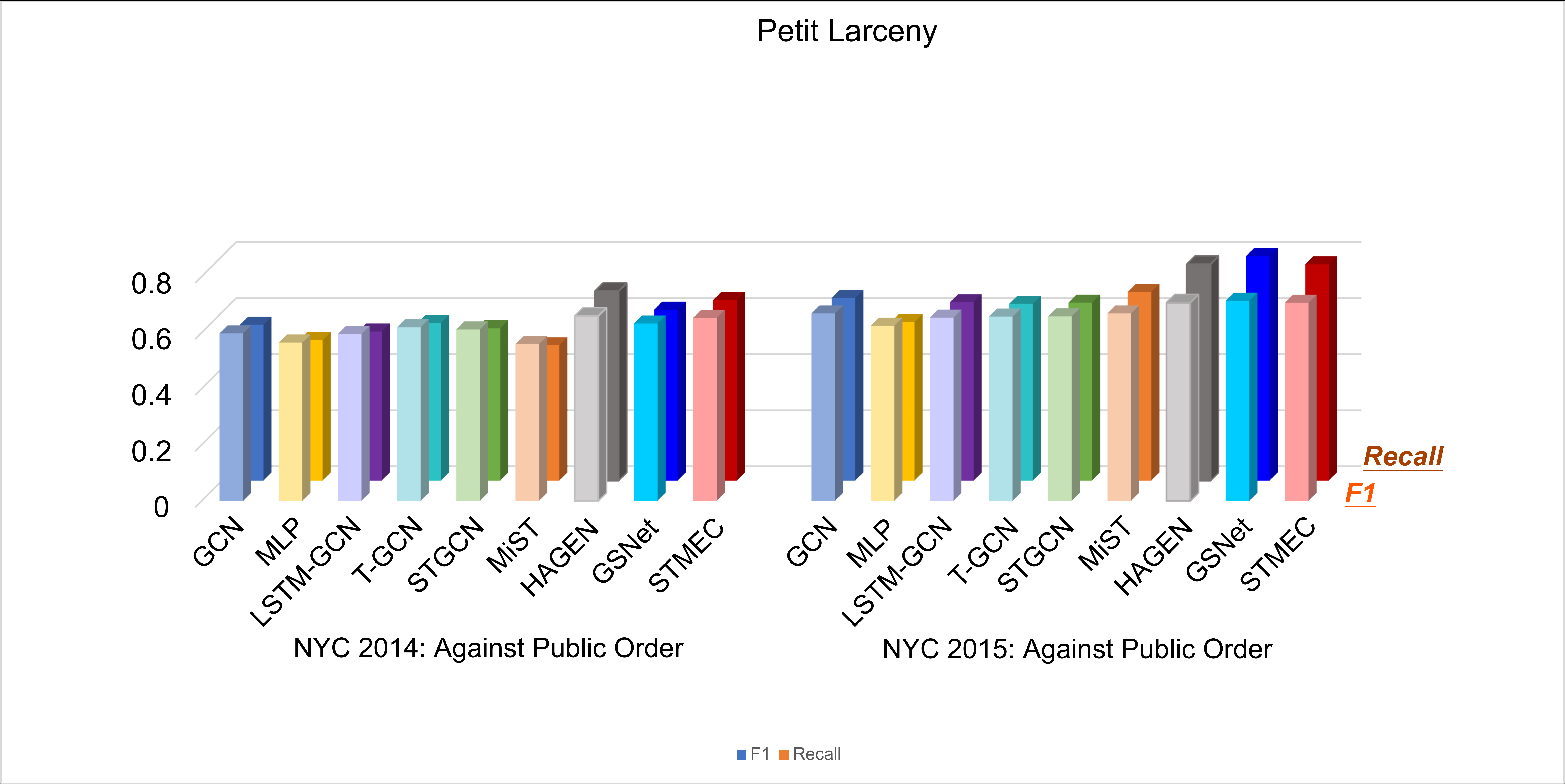}
         \caption{Against Public Order}
     \end{subfigure}
     \hfill
     \begin{subfigure}[b]{0.49\textwidth}
         \centering
         \includegraphics[width=\textwidth]{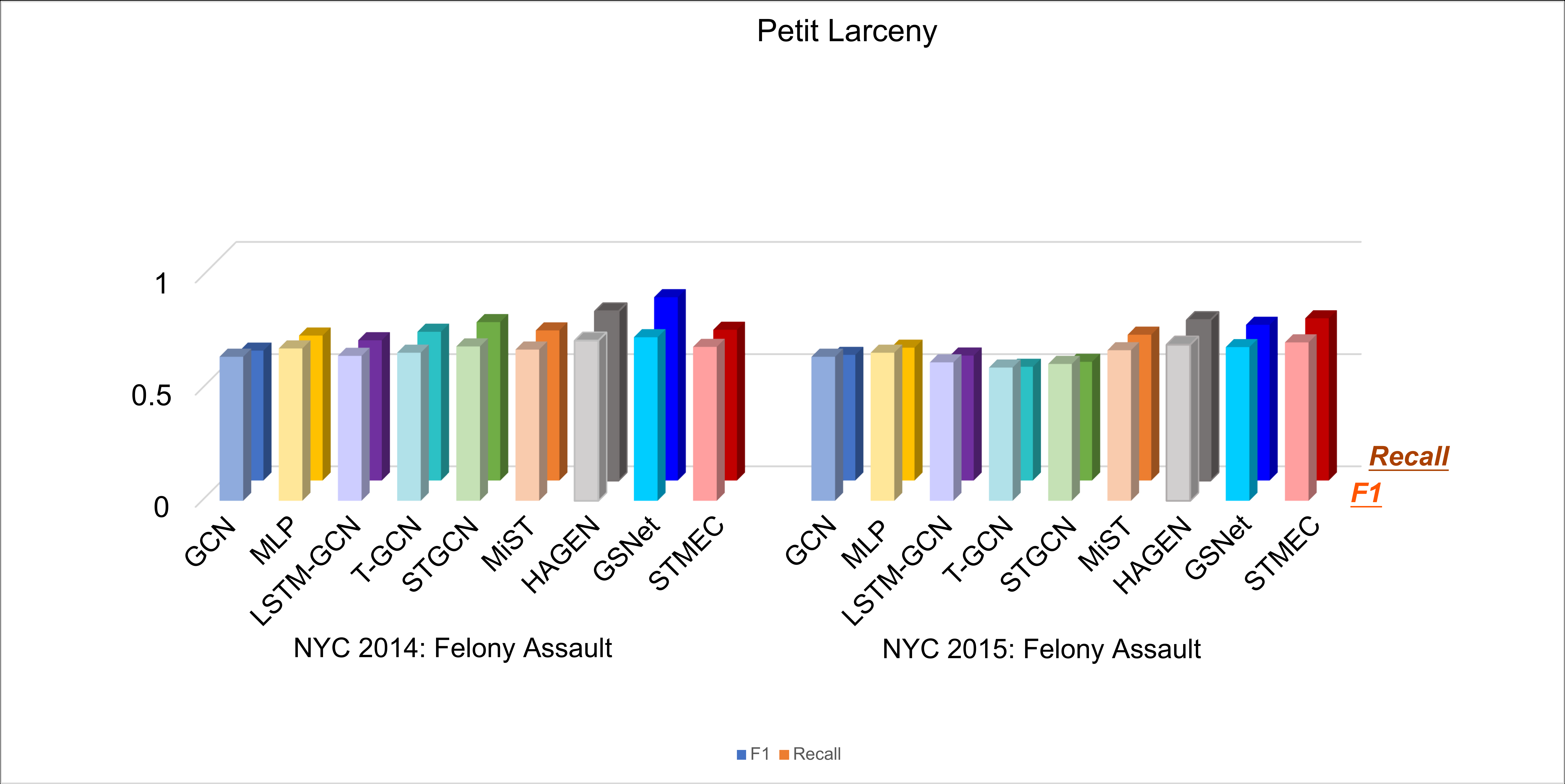}
         \caption{Felony Assault}
     \end{subfigure}
     \hfill
     \begin{subfigure}[b]{0.49\textwidth}
         \centering
         \includegraphics[width=\textwidth]{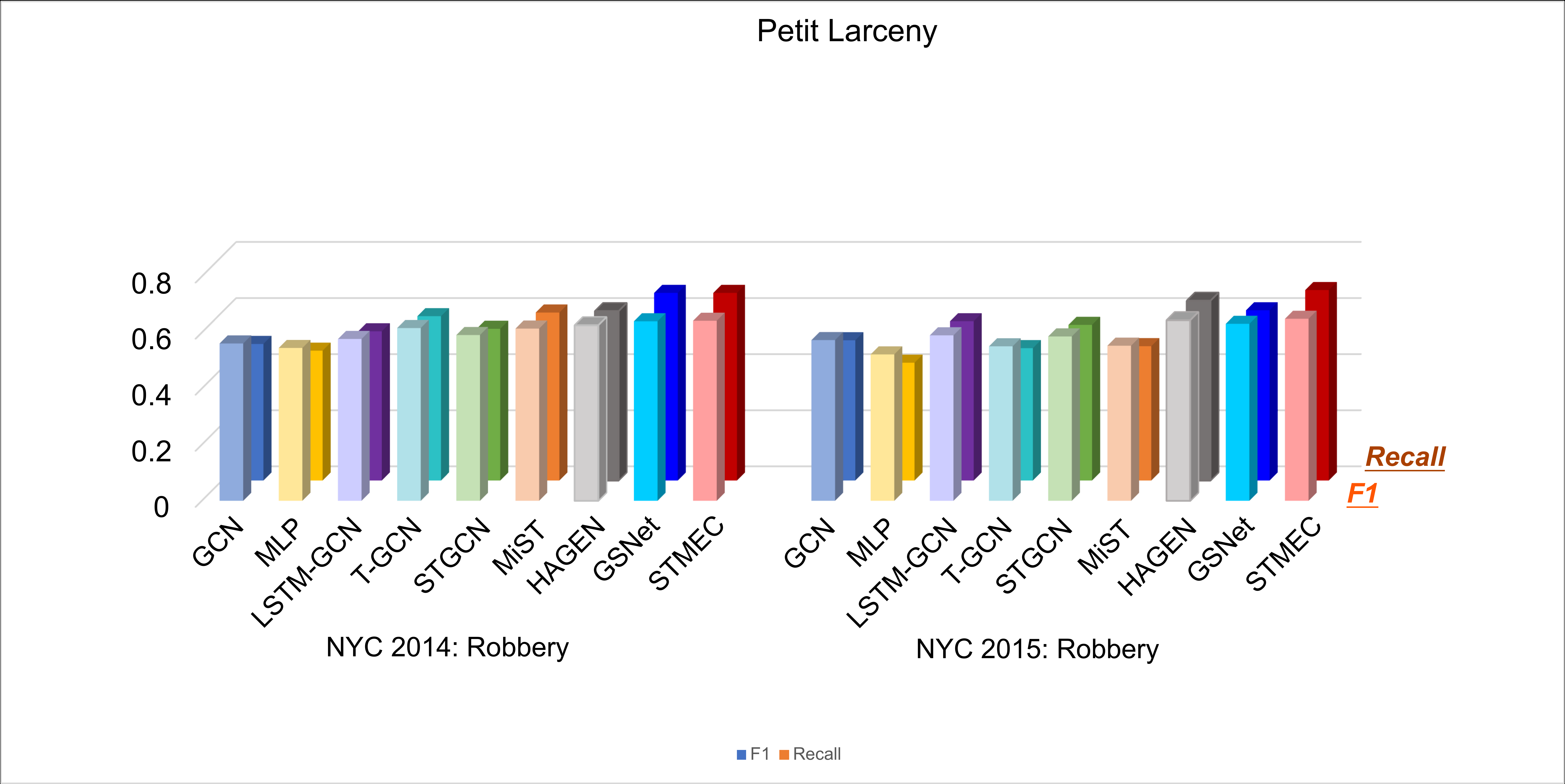}
         \caption{Robbery}
     \end{subfigure}
     \hfill
     \begin{subfigure}[b]{0.49\textwidth}
         \centering
         \includegraphics[width=\textwidth]{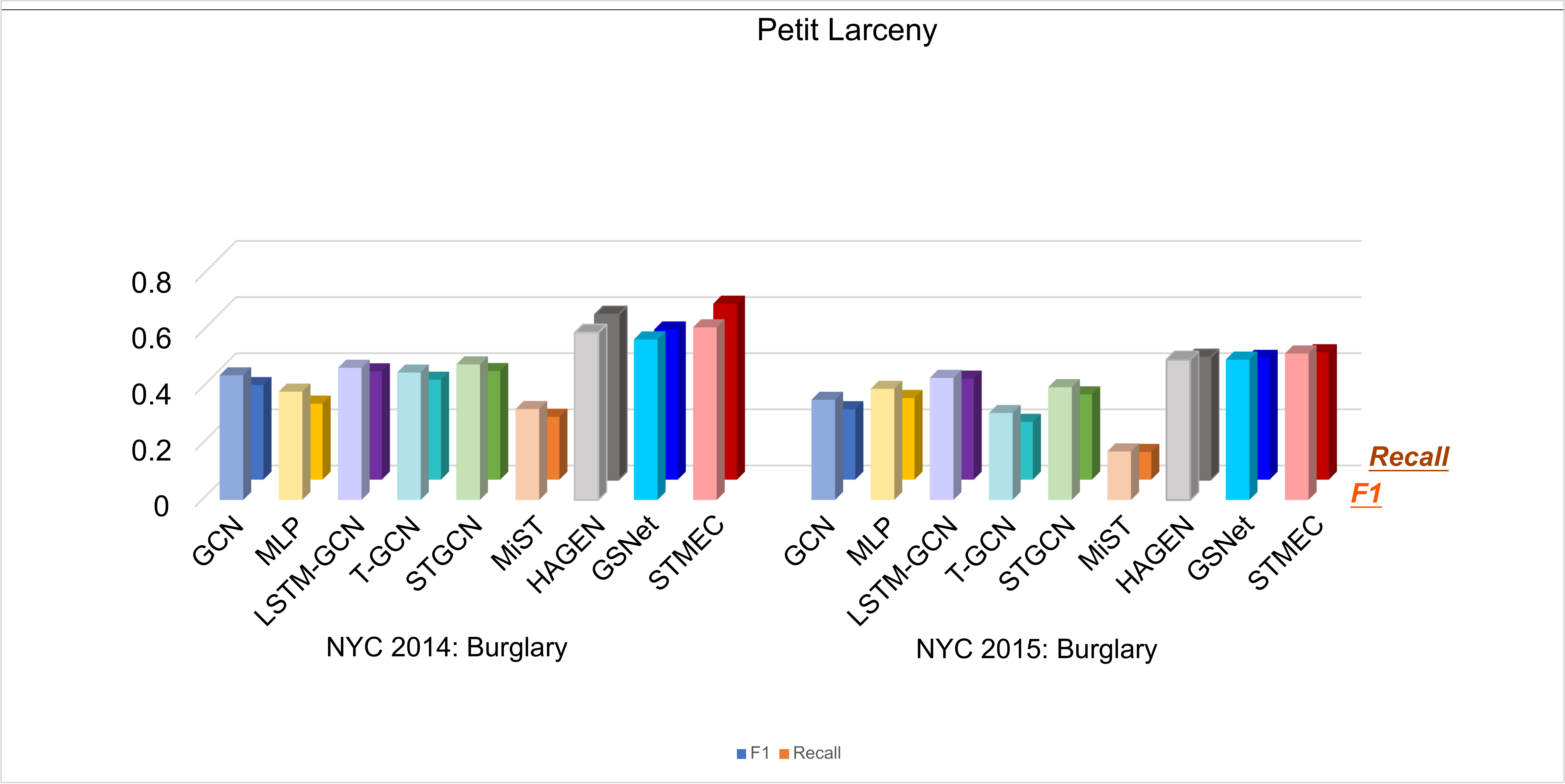}
         \caption{Burglary}
     \end{subfigure}
     \hfill
     \begin{subfigure}[b]{0.49\textwidth}
         \centering
         \includegraphics[width=\textwidth]{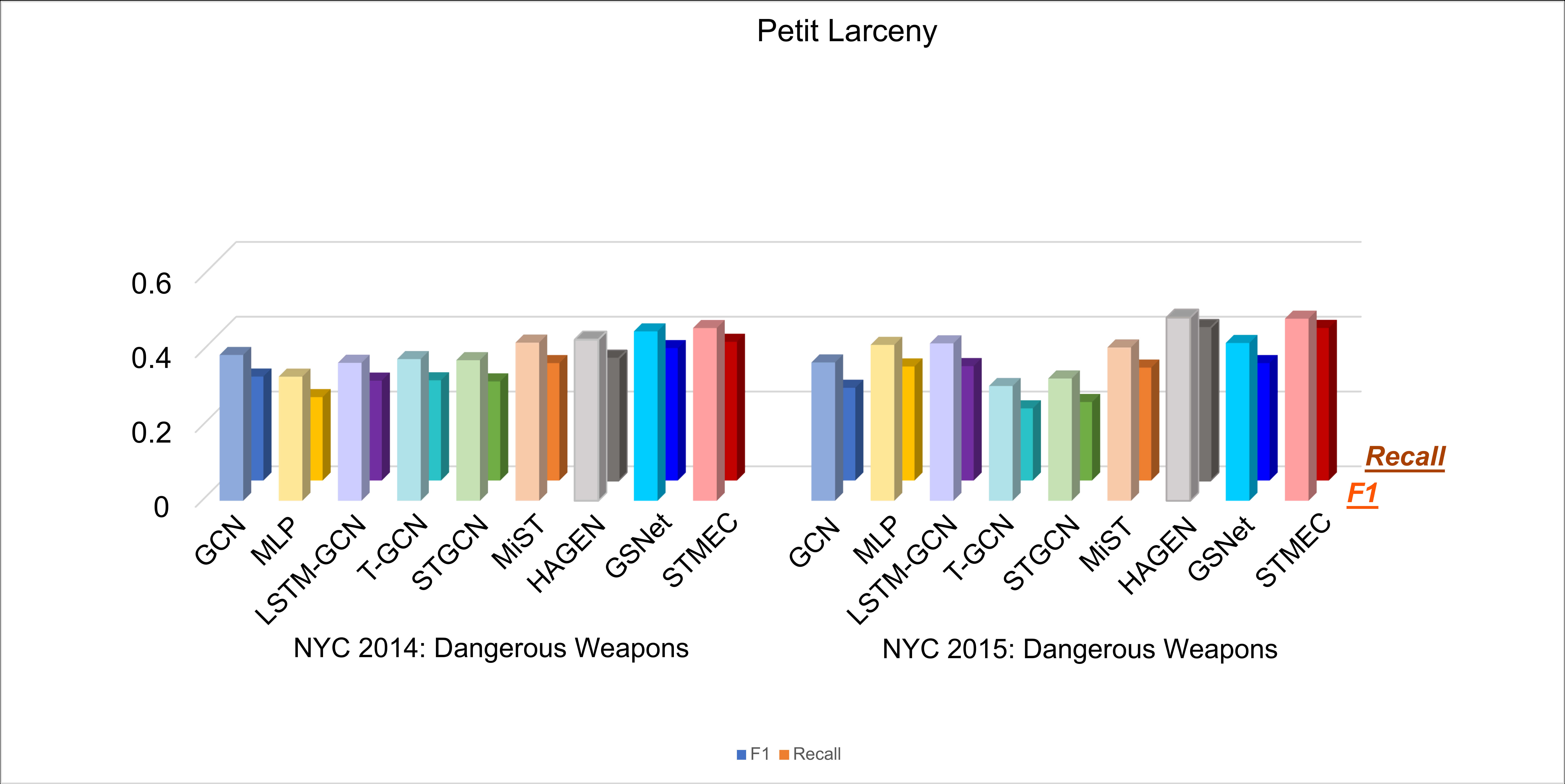}
         \caption{Dangerous Weapons}
     \end{subfigure}
     \hfill
     \begin{subfigure}[b]{0.49\textwidth}
         \centering
         \includegraphics[width=\textwidth]{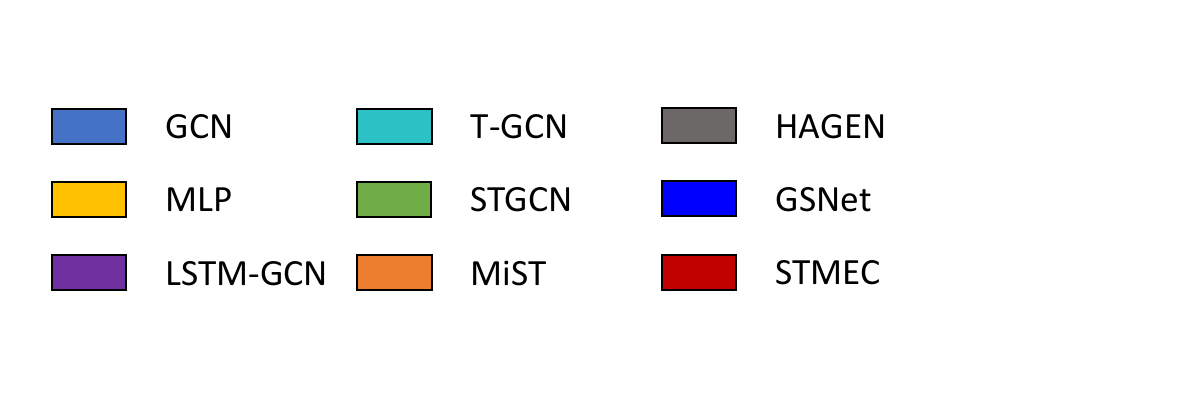}
     \end{subfigure}
     \hfill
        \caption{{Performance comparison by each type of crime events.}}
        \label{fig:comparison}
\end{figure}

Additionally, we explore the effectiveness of STMEC in forecasting each category of crime events. From Figure \ref{fig:comparison}, we observe that for frequent crime events (from \textbf{(a) Petit Larceny} to \textbf{(e) Grand Larceny}), most of the deep learning baselines can persist remarkable performances while STMEC stays competitive for predicting those kinds of crimes. However, for crime events that rarely happen (from \textbf{(f) Dangerous Drugs} to \textbf{(k) Dangerous Weapons}), most of the baselines fall to capture the dynamic patterns and undermine the ability to identify the occurrence of criminal activities. {Although STMEC falls slightly behind HAGEN and GSNet in predicting crimes related to against public order and felony assault, it alleviates the scarcity of other rarer crime types by achieving 5\%-20\% higher recall and 3\%-10\% higher F1 score. For instance, when predicting dangerous drugs related crimes in 2015, STMEC achieves recall of $0.744$ and F1 score of $0.72$. It outperforms the second-best baseline model GSNet which achieves recall of $0.644$ and F1 score of $0.685$.} Furthermore, STMEC achieves the highest recall and F1 score of predicting burglary in 2014, which are $0.627$ and $0.615$ respectively. {This result reveals 6.7\% and 4.3\% improvement in terms of recall and F1 score compared to the best baseline HAGEN.} Given the significant improvement in predicting uncommon crimes, we can conclude that the performance of STMEC is stable and robust across different crime types.

\clearpage

\begin{table}
\renewcommand{\arraystretch}{1.5}
\centering
\footnotesize
   \begin{tabular}{M{0.15\textwidth}M{0.13\textwidth}M{0.13\textwidth}M{0.13\textwidth}M{0.13\textwidth}}
    \toprule
    \textbf{Factors} & \textbf{Mean} & \textbf{Min.} & \textbf{Median} & \textbf{Max.}  \\
    \midrule
    Demographics & 0.167 & 0.049 & 0.157 & 0.359\\
    \midrule
    Income & 0.177 & 0.104 & 0.164 & 0.375\\
    \midrule
    Job & 0.00019 & 0.00012 & 0.00018 & 0.00027\\
    \midrule
    Commuting & 0.112 & 0.004 & 0.085 & 0.459\\
    \midrule
    Urbanization & 0.502 & 0.167 & 0.513 & 0.744\\
    \midrule
    Geographic & 0.040 & 0.012 & 0.037 & 0.078\\
    \bottomrule
\end{tabular}
\caption{Summary table of importance score obtained from attention.}
 \label{tab:summary}
\end{table}

\begin{figure}[ht]
  \centering
  \includegraphics[width=\linewidth]{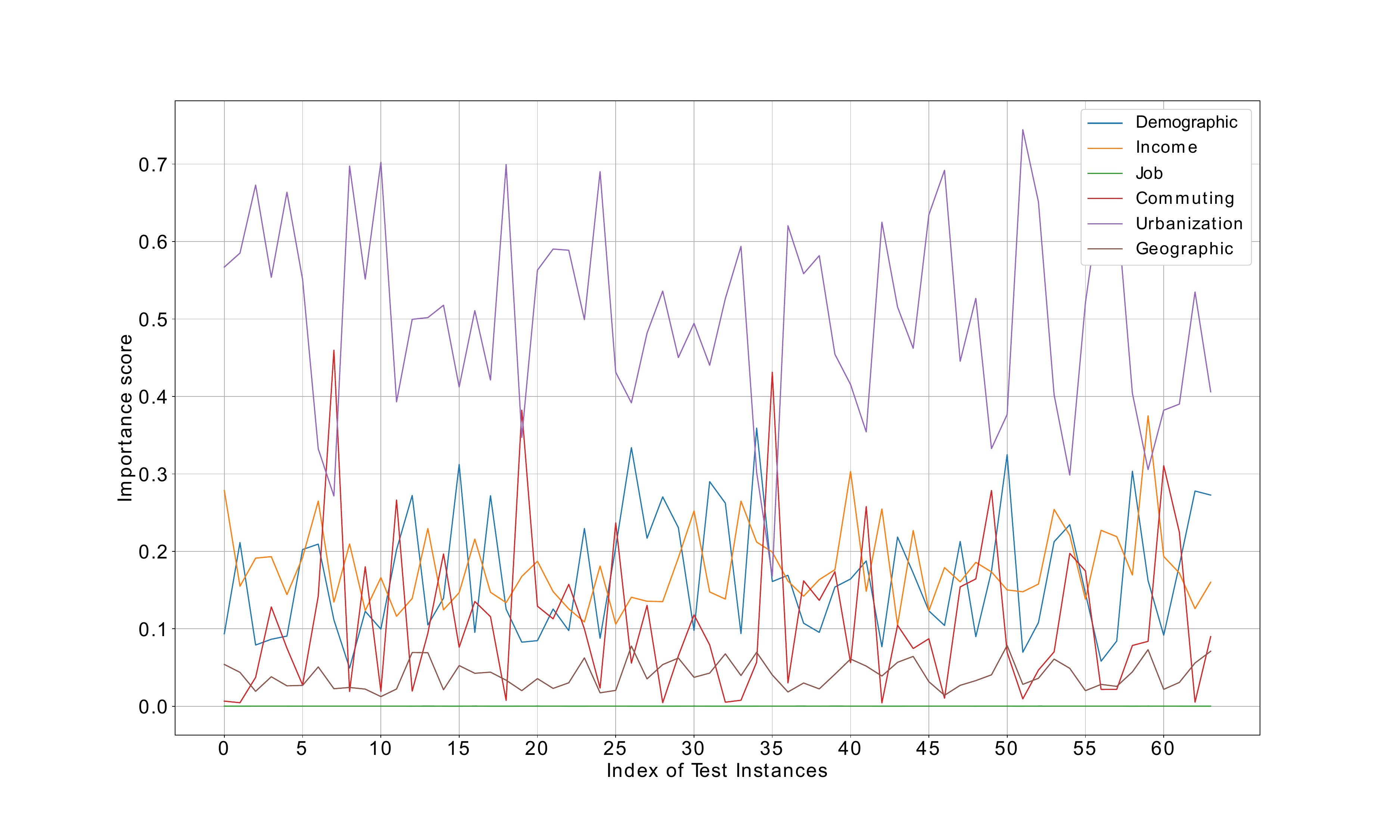}
  \caption{Variations of attention scores in testset (NYC 2014).}
  \label{fig:scores}
\end{figure}

\subsection{Analysis of Explainability}\label{subsec11}
Apart from the superior performance given by STMEC, another contribution of the framework is to give explanations on forecasts. This work involves 6 different types of meta-paths which represent the features from different categories of potential factors. As we generate pre-defined region-to-region instances and sum up the impacts from each path by a given similarity score, the effect of the black box is relieved by explicitly modeling the interactions and ranking priorities of the path instances. Furthermore, the attention mechanism that obtains the weight of each meta-path type also helps to rank the importance of factors in the crime prediction tasks. Table \ref{tab:summary} demonstrates the summary of importance sores with respect to different factors. As the scores are obtained from each test instance given by day, the result indicates that the importance of each factor varies a lot along the timeline. Specifically, we can observe the variation of importance for each factor from Figure \ref{fig:scores}. It shows that urbanization level is the leading factor for most of our predictions, while the type of job in the population is the least important factor in the task. Additionally, we can also observe that the impact of geographical information is relatively stable, while the others dramatically change up and down. It is worth noticing that the causes of criminal activities are not constant along the time. This observation supports our idea of constructing the model by considering the variations and interactions between features.

\begin{figure}[ht]
    \centering
     \begin{subfigure}[b]{\textwidth}
         \centering
         \includegraphics[width=\textwidth]{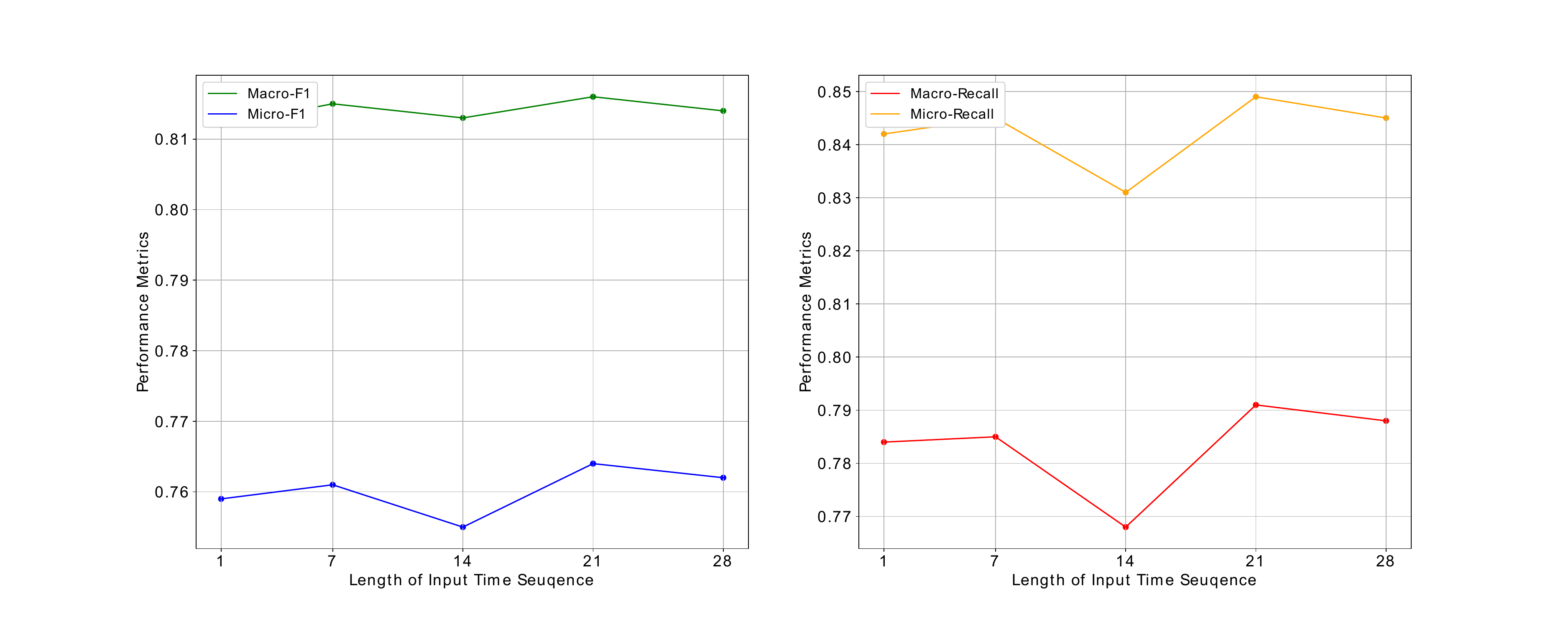}
         \caption{NYC 2014.}
     \end{subfigure}
     \hfill
     \begin{subfigure}[b]{0.85\linewidth}
         \centering
         \includegraphics[width=\textwidth]{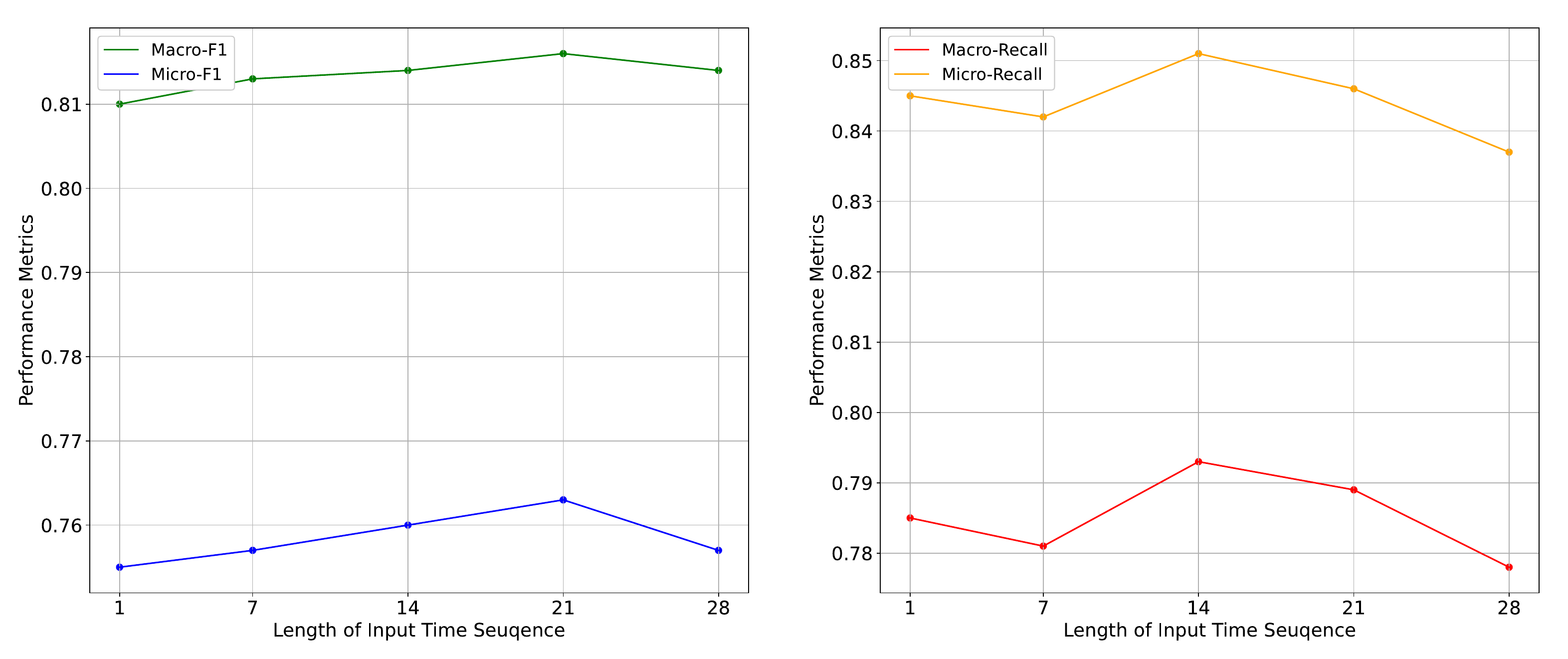}
         \caption{{NYC 2015.}}
         \label{fig:time2015}
     \end{subfigure}
\caption{{Performance of model across different time window sizes.}}
        \label{fig:time}
\end{figure}

\subsection{Hyperparameter Study}\label{subsec12}
\subsubsection{Effect of the Temporal Scale}\label{time_w}
\begin{figure}[ht]
    \centering
     \begin{subfigure}[b]{0.85\textwidth}
         \centering
         \includegraphics[width=\textwidth]{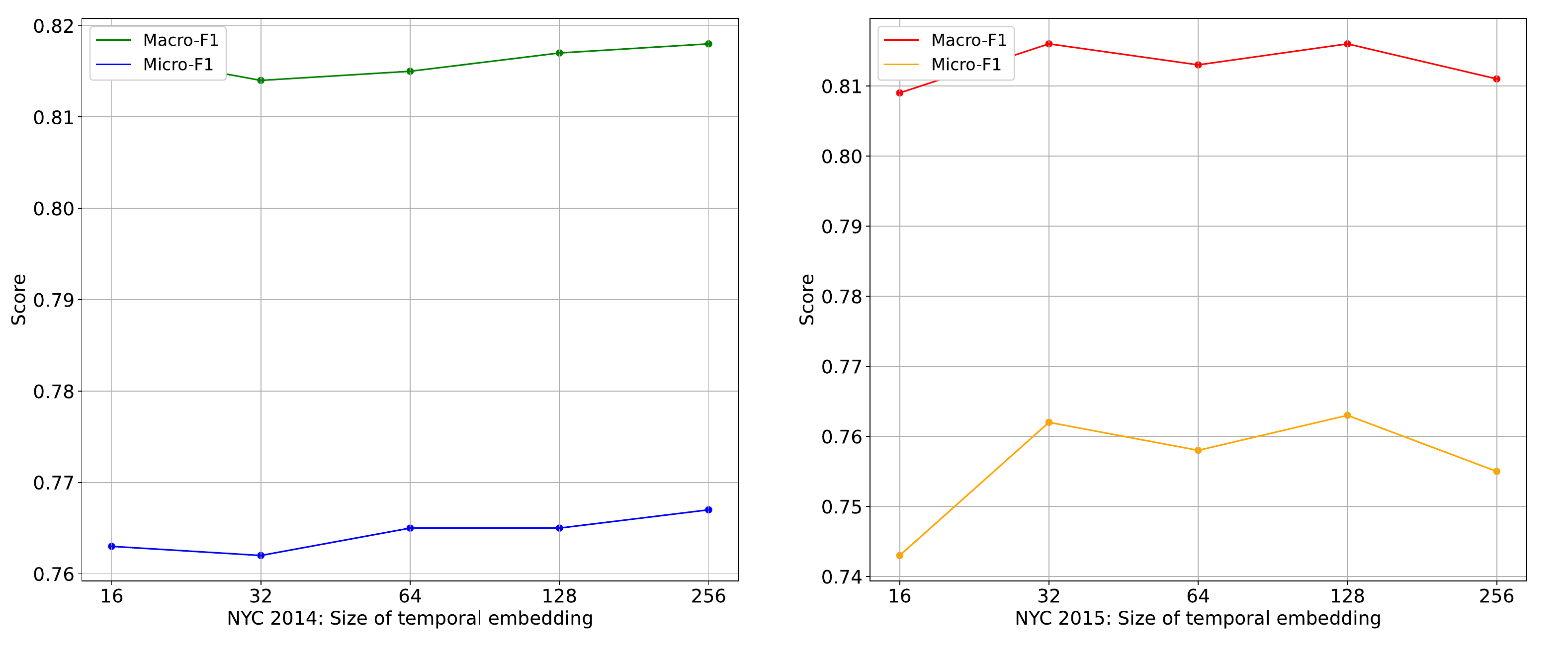}
         \caption{Impact of temporal embedding.}
         \label{fig:te}
     \end{subfigure}
     \hfill
     \begin{subfigure}[b]{0.85\linewidth}
         \centering
         \includegraphics[width=\textwidth]{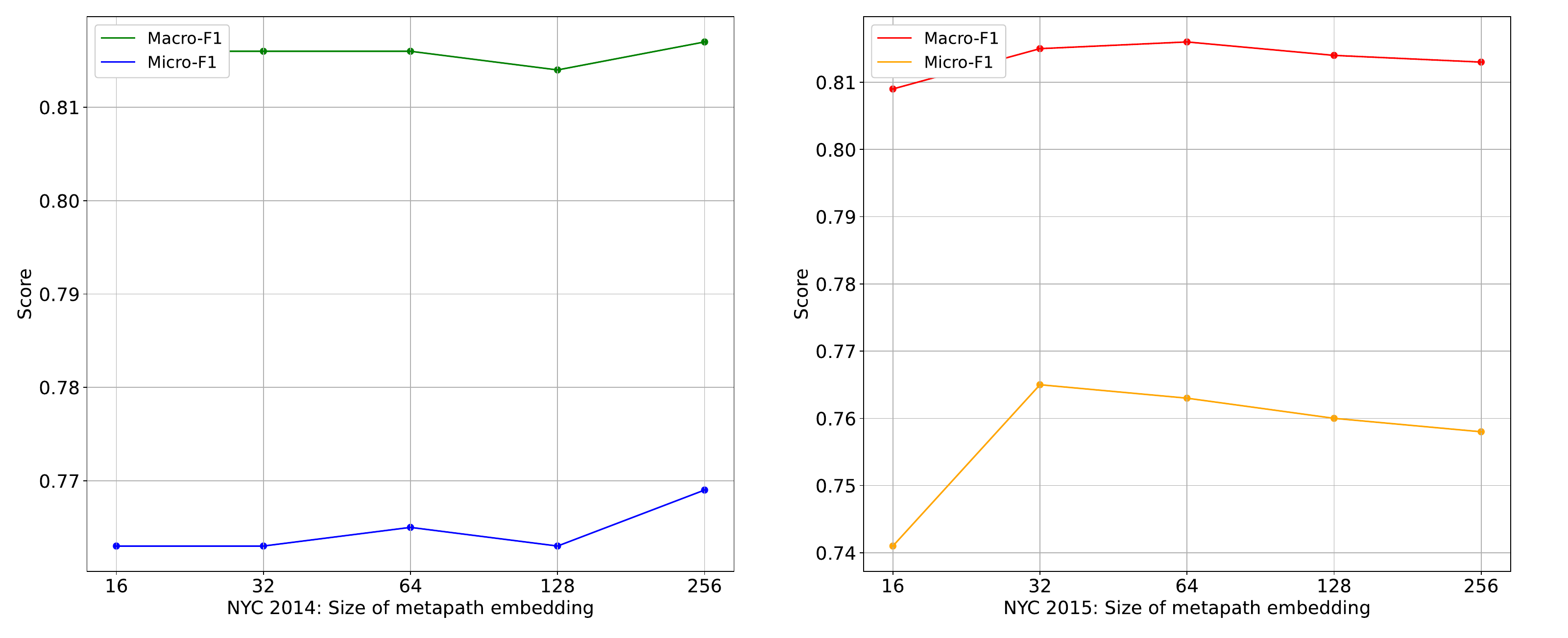}
         \caption{Impact of meta-path embedding.}
         \label{fig:me}
     \end{subfigure}
\caption{Impact of embedding size.}
        \label{fig:eb}
\end{figure}

Since the crime prediction is conducted by utilizing the last $M$ days of Observations, the length of the encoded time-series sequence is one of the most essential parameters to assist in capturing the temporal dynamics. However, as we observed in Section 4.3.1, the experimental results reveal that the difference between encoding short-term time dependency (one day) and long-term time dependency (up to 4 weeks) is not significant. To further justify this observation and choose the optimal size of the time window for our proposed model, we investigate the change of performance across different length of input sequence $M$. Figure \ref{fig:time} depicts the performance of STMEC changes across different choice of time window $M$. As we can observe from the figure, for dataset NYC 2014, the performance of STMEC becomes better as the size of the time window $M$ increases, but saturates when $M$ reaches 21. {Also, as we further analyze the impact of time window for NYC 2015 in Figure \ref{fig:time2015}, the experimental results suggest that the optimal window size that achieves the highest F1 score is still 21, but the model obtains better recall when we set the window size to 14. Thus, based
on different patterns shown in different datasets, we can infer that the change of performance under different time window settings is mainly due to the specific periodicity of crime trends in the datasets.} The result is consistent with the observation in Section \ref{subsubsec4}. \textcolor{black}{As the experimental results in Section \ref{subsubsec4} suggest, the crime trends of NYC 2015 tend to have more temporal patterns in the short term, which benefits MLP and GCN that only use features from one step earlier and weakens TGCN that will possibly capture the noise from long-term data. The ACF plots in Fig.\ref{fig:acf} support our observation, as the recent lags of crime trends in 2015 have larger correlation values compared with crime trends in 2014.} By choosing the optimal size of the time window, the performance of the model is slightly improved even compared to the worst option that set the size of the time window to 1. Hence, the crime events are not as temporal-correlated as we thought, and we can barely improve the crime prediction tasks more by solely investigating the temporal features.

{\subsubsection{Effect of the Embedding Size}\label{subsec13}
To understand the effect of different embedding size, we further examine the size of temporal embedding in LSTM component, and the size of meta-path embedding during node aggregation. As shown in Figure \ref{fig:te} and Figure \ref{fig:me}, when the size of temporal embedding is set to 128 and the size of meta-path embedding is set to 64, the model can achieve slightly better F1 scores as compared to the others. However, as the difference between F1 scores is only around 1\%, we can conclude that our model is insensitive to the size of latent representations.}

\begin{figure}[ht]
    \centering
     \begin{subfigure}[b]{0.7\textwidth}
         \centering
         \includegraphics[width=\textwidth]{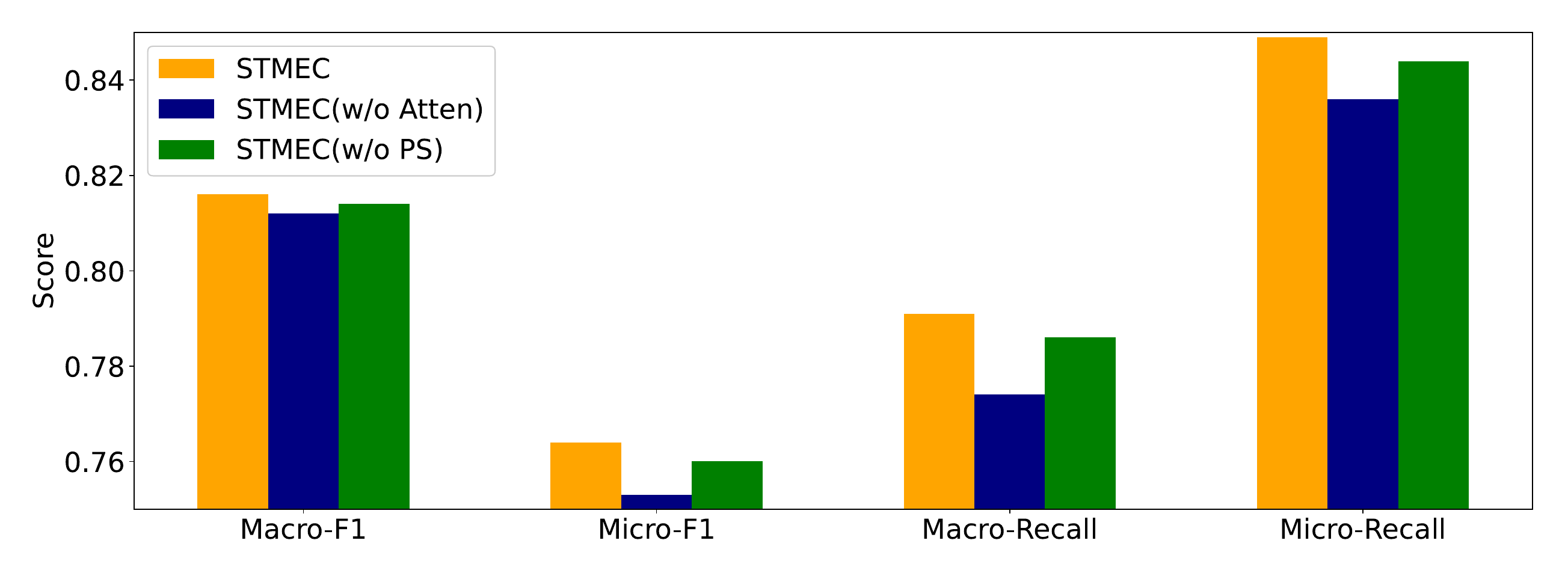}
         \caption{NYC 2014.}
         \label{fig:as14}
     \end{subfigure}
     \hfill
     \begin{subfigure}[b]{0.7\linewidth}
         \centering
         \includegraphics[width=\textwidth]{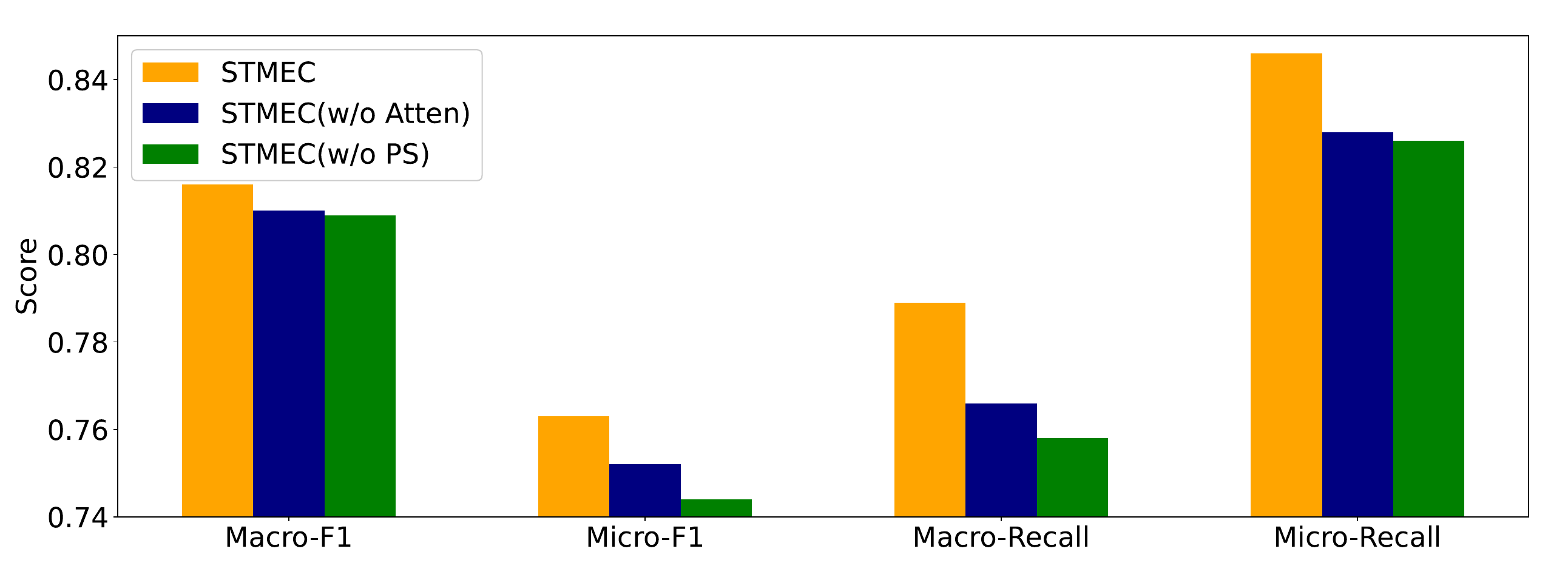}
         \caption{NYC 2015.}
         \label{fig:as15}
     \end{subfigure}
    \caption{Significance of different component in STMEC.}
        \label{fig:as}
\end{figure}

{
\subsection{Ablation Study}
To understand the significance of different components in our proposed framework, we further examine the efficacy of the attention module and the distribution-aware PathSim by replacing them with a mean aggregator in the degraded variants of STMEC. In general, the experimental results in Figure \ref{fig:as} have shown that both key components are beneficial to the performance of our model. Besides, attention module reflects more significance on predicting crimes of NYC 2014, while PathSim shows more potential in predicting crimes of NYC 2015.}

\section{Related Work}\label{sec5}
\textbf{Urban mobility dynamics}. Apart from investigating ubiquitous data and historical crime records over time, some studies also analyze the influence of urban mobility dynamics on crime events \cite{Guangwen01, 10.1145/3274895.3274994, 10.1145/3340531.3412065, Wang2020DeepTM, Cristina01, Raquel01}. \cite{10.1145/3274895.3274994} proposed the User Associated Dynamic Crime Risk (UADCR) features to associate human movement and the risk of a region that may involve in crime events. Based on this work, \cite{10.1145/3340531.3412065} built a directed weighted graph to analyze the relationship between crime rate and human movements in different periods of the day. Apart from this, \cite{Cristina01} also crafts the spatial-temporal features of human activities from social networks and transportation data, and compared the performance between conventional methods and the tree based machine learning techniques.

\textbf{Semantic representation learning}. As spatial and temporal patterns from human mobility reveal the underlying causes of criminal behaviors, the semantics behind the activity patterns are also discussed in some work \cite{10.1145/3459637.3482482, 10.1145/3158433}. Semantic representation learning can strengthen the understanding of the urban dynamics and issues in various tasks, such as crime, traffic, and user demographics in urban regions. SUME \cite{ 10.1145/3411807} learns a semantic-enhanced embedding of the heterogeneous network which has various types of nodes and edges to infer user demographics. Although this framework is set to optimize user profiles, the implicit representation of urban regions is also learned throughout the task, which sheds light on the possibilities that such similarity based measurements can also help deal with region-level crime prediction tasks. Regarding the semantic learning in crime predictions, one most recent work \cite{10.1145/3459637.3482482} handles dynamic chain effect and multidimensional features for performing multi-incident co-prediction. This work explicitly exploits the behind-the-scenes chain-like triggering mechanism and tackles the challenging problem caused by incident sparsity. {Further more, to learn embeddings from social networks, LBSN2Vec++\cite{yang2020lbsn2vec} is proposed to preserve the information of social relationships and user mobility by encoding friendship edges between users and category hyperedges across different node domains. } 

\textbf{Spatial-temporal modelling}. There are also plenty of recent studies that mine the spatial-temporal dependencies for other stochastic systems, such as human trajectory \cite{10.1109/ICCV.2019.00637, 10.1145/3340531.3412054}, e-commerce \cite{chen2018tada,chen2020sequence}, social network \cite{tam2019anomaly,trung2020adaptive} and water quality \cite{10.1145/3308558.3313577}. Most of the recent methods are deep learning based models and well capture spatial-temporal dependencies with respect to the particularity of certain tasks. For example, \cite{10.1109/ICCV.2019.00637} proposes a graph attention based sequence-to-sequence model to predict crowd trajectories. Considering that human motion is continuous and forward-looking, the framework explicitly models the continuity of interactions among pedestrians. Additionally, \cite{10.1145/3308558.3313577} utilizes transfer learning to aggregate information from multiple cities and captures long-term periodicity by a pattern based spatial-temporal memory. {Further more, the traffic accident prediction framework GSNet \cite{wang2021gsnet} models different kinds of spatial correlations by different types of context features. However, using separate homogeneous graphs inevitably incurs sparse connections between region nodes as there is only one relation type considered. Different to existing spatial-temporal models, our proposed method can learn high-quality region embeddings by utilizing meta-paths in a unified HIN, which allows for simultaneously learning the importance of different factors towards the prediction results.}

{\textbf{Crime prediction}. Although most of the existing work aims to predict the occurrence of criminal activities, the analysis of more specific factors (e.g., income level) is left untouched in deep learning based models. Unlike traditional regression models where the coefficients directly indicate the importance of features \cite{wang2016crime}, it is still a challenge to explain the feature importance in state-of-the-art deep methods for crime prediction. Also, while some work \cite{wang2016crime, wang2013learning, yang2018crimetelescope} has widely explored the demographics, POIs, and geographical features in crime prediction tasks, our model novelly depicts the relations between regions by different socioeconomic factors, and its performance and explainability are strengthened via the path-enriched features in the graph-structured data. In addition to exploring spatial-temporal impacts on criminal events, some studies \cite{zhao2022multi, 10.1145/3308558.3313730, wang2022hagen} also include the cross-type correlations of urban crimes to enhance the accuracy of crime prediction. CCC \cite{zhao2022multi} jointly captures the intrinsic correlations between crime types and the spatial-temporal correlations of criminal activities by mathematical modeling. As our proposed model focus more on the impact of different socioeconomic factors on the occurrence of different crimes, there is potential to further capture correlations between different crime types in our future work. }

\section{Conclusion}\label{sec6}
In this paper, we propose a novel framework STMEC which explicitly models the dynamic interactions between regions over time with knowledge-aware paths from multiple views. We also explore the contribution of different factors that potentially result in criminal activities with attention mechanism. The experiments from two real-world crime events datasets show that the proposed framework outperforms state-of-the-art baselines on key metrics of performance and explainability. Future work may include the following directions: 1) Integrate the impact of urban mobility/human activities, and 2) Explore the importance score of different factors of each specific region.

\section{Statements and Declarations}\label{sec8}
\subsection{Ethical Approval and Consent to participate}\label{}
This work has been reviewed by the Research Ethics and Integrity and is deemed to be exempt from ethics review under the National Statement on Ethical Conduct in Human Research and relevant University of Queensland policy (PPL 4.20.07).

\subsection{Human and Animal Ethics}\label{}
This article does not contain any experimentation with human or animal subjects.

\subsection{Consent for Publication}\label{}
Not applicable.

\subsection{Availability of Supporting Data}\label{}
All datasets used for supporting the conclusions of this article are available from the public data repository at the website of data.cityofnewyork.us and www.kaggle.com. 

\subsection{Competing Interests}\label{}
All authors have no competing interests as defined by Springer, or other interests that might be perceived to influence the results and/or discussion reported in this paper.

\subsection{Funding}\label{}
No funding was received for conducting this study.

\subsection{Authors' Contributions}\label{}
Yuting Sun implemented the methodology, performed data curation, wrote the original draft, and conducted the experiment.

Tong Chen proposed the conceptualization and methodology, prepared the figures, and performed review and editing. 

Hongzhi Yin proposed the research problem and methodology, supervised the whole research work, wrote the introduction and abstract parts, and performed review and editing.

\subsection{Acknowledgements}\label{}
We gratefully thank Dr. Thomas Taimre (the University of Queensland) and Dr. Radislav Vaisman (the University of Queensland) for valuable discussions on this research and helpful comments on the manuscript. This work is supported by Australian Research Council Future Fellowship (Grant No. FT210100624) and Discovery Project (Grant No.DP190101985).

\subsection{Authors' Information}\label{}
Yuting Sun, Ph.d candidature, School of Information Technology and Electrical Engineering, the University of Queesnland, Brisbane, Australia.

Tong Chen, Lecturer, School of Information Technology and Electrical Engineering, the University of Queesnland, Brisbane, Australia.

Hongzhi Yin, Associate Professor/ARC Future Fellow, School of Information Technology and Electrical Engineering, the University of Queesnland, Brisbane, Australia.







\bibliography{citation}


\end{document}